\documentclass[conf]{new-aiaa}

\usepackage[utf8]{inputenc}

\usepackage{graphicx}
\usepackage{amsmath}
\usepackage[version=4]{mhchem}
\usepackage{siunitx}
\usepackage{longtable,tabularx}
\usepackage{multirow}
\usepackage[table,xcdraw]{xcolor}
\usepackage{algorithm}
\usepackage{algorithmic}
\usepackage{booktabs}
\usepackage[table,xcdraw]{xcolor}
\renewcommand{\arraystretch}{1.5}

\hypersetup{
	colorlinks   = true, %Colours links instead of ugly boxes
	urlcolor     = blue, %Colour for external hyperlinks
	linkcolor    = blue, %Colour of internal links
	citecolor    = blue %Colour of citations
}

\title{A Multi-Fidelity Methodology for Reduced Order Models with High-Dimensional Inputs}

\author{Bilal Mufti\footnote{Graduate Researcher, ASDL, Daniel Guggenheim School of Aerospace Engineering}}

\author{Christian Perron\footnote{Research Engineer II, ASDL, Daniel Guggenheim School of Aerospace Engineering}}

\author{Dimitri N. Mavris\footnote{S.P. Langley Distinguished Regents Professor and Director of ASDL, Daniel Guggenheim School of Aerospace Engineering,  AIAA Fellow}}
\affil{Georgia Institute of Technology, Atlanta, Georgia, 30332}

\begin{document}

\maketitle

\begin{abstract}
   In the early stages of aerospace design, reduced order models (ROMs) are crucial for minimizing computational costs associated with using physics-rich field information in many-query scenarios requiring multiple evaluations. The intricacy of aerospace design demands the use of high-dimensional design spaces to capture detailed features and design variability accurately. However, these spaces introduce significant challenges, including the curse of dimensionality, which stems from both high-dimensional inputs and outputs necessitating substantial training data and computational effort. To address these complexities, this study introduces a novel multi-fidelity, parametric, and non-intrusive ROM framework designed for high-dimensional contexts. It integrates machine learning techniques for manifold alignment and dimension reduction—employing Proper Orthogonal Decomposition (POD) and Model-based Active Subspace—with multi-fidelity regression for ROM construction. Our approach is validated through two test cases: the 2D RAE~2822 airfoil and the 3D NASA CRM wing, assessing combinations of various fidelity levels, training data ratios, and sample sizes. Compared to the single-fidelity PCAS method, our multi-fidelity solution offers improved cost-accuracy benefits and achieves better predictive accuracy with reduced computational demands. Moreover, our methodology outperforms the manifold-aligned ROM (MA-ROM) method by 50\% in handling scenarios with large input dimensions, underscoring its efficacy in addressing the complex challenges of aerospace design.
\end{abstract}

\section*{Nomenclature}

{\renewcommand\arraystretch{1.0}
\noindent

\begin{longtable*}{@{}l @{\quad=\quad} l@{}}
    $\mathbf{Y}, \mathbf{Z}$  & high- and low-fidelity training fields \\
    $\mathcal{X}$ & input parameter space \\
    $Re_{\infty}$ & free stream Reynolds number \\
    $M_{\infty}$ & free stream Mach number \\
    $C_l, C_d$ & lift and drag coefficients \\
    $C_p$ & coefficient of pressure \\
    $m_1, m_2$ & number of high- and low-fidelity samples \\
    $p, q$ & dimensionality of high- and low-fidelity fields\\
    $d$ & input space dimensionality \\
    $n_{v}$ & number of test samples \\
    $\tau$ & multi-fidelity ratio parameter \\
    $\tilde{E}, \tilde{E}_{RC}, \tilde{E}_{RG}$ & field prediction, reconstruction, and regression errors \\
\end{longtable*}
\noindent
\emph{Procrustes Manifold Alignment}
\begin{longtable*}{@{}l @{\quad=\quad} l@{}}
    $\mathbf{H}, \mathbf{S}$  & high- and low-fidelity field latent coordinates \\
    $k$ & latent output space dimensionality \\
    $\mathbf{\Theta}_{k}, \mathbf{\Xi}_{k}$  & high- and low-fidelity POD modes \\
    $\mathbf{U}, \mathbf{V}$ & squares matrices containing left and right singular vectors \\
    $\boldsymbol{\Sigma}$ & diagonal matrix of singular values \\
    $t$ & scaling factor\\
    $\mathbf{G}$ & low-fidelity latent variables aligned with the high-fidelity variables
\end{longtable*}

\noindent
\emph{Model-Based Active Subspace}
\begin{longtable*}{@{}l @{\quad=\quad} l@{}}
    $\delta$  & correction term for the difference between high- and low-fidelity latent variables\\
    $a, \mathbf{b}$  & linear regression coefficients \\
    $w$ & weights of surrogate model\\
    $\phi,\epsilon$ & basis function and polynomial trend \\
    $\mathbf{C}_{MF}$ & multi-fidelity covariance matrix of gradients \\
    $\mathbf{\Lambda}$ & diagonal matrix of eigenvalues \\
    $\mathbf{W}$ & orthogonal matrix containing eigenvectors \\
    $\mathbf{W}_{AS}$ & active subspace matrix \\
\end{longtable*}

\noindent
\emph{Subscripts and Superscripts}
\begin{longtable*}{@{}l @{\quad=\quad} l@{}}
    $L, U$  & linked and unlinked data \\
    $\Tilde{}$  & model prediction\\
    $*$ & out-of-sample design \\
    $AS$ & active subspace \\
    $\mathbf{'}$ & gradient of the function \\
\end{longtable*}
}

\section{Introduction}

The last few decades have witnessed an exponential growth in global computational power. However, this increase has not significantly offset the computational cost of certain high-fidelity numerical simulations, particularly those employing discretization methods like finite element or finite volume \cite{benner2015survey}. These simulations become resource-intensive in \emph{many-query} scenarios such as design space exploration, design optimization, uncertainty quantification, and propagation~\cite{yondo2018review}. In aerospace design, the early design phase frequently uses these high-fidelity simulations in many-query contexts to assess multiple design configurations and to conduct in-depth multi-disciplinary analyses. Yet, the high computational expense of these simulations presents a substantial challenge to their practical application in the initial stages of aerospace systems design. This situation has driven designers to seek and adopt more efficient methods and techniques that can alleviate the computational load without compromising the quality and depth of the analyses.

 Reduced order modeling is emerging as a powerful tool in the engineering and computational science fields due to its ability to significantly reduce the computational cost of obtaining high-fidelity numerical simulation results. Reduced order models (ROMs) serve a purpose similar to traditional surrogate models, i.e., they offer a cheap approximation of an expensive numerical simulation. However, unlike traditional data-fit surrogate models that are used to predict integrated scalar quantities, ROMs can be used to represent high-dimensional field solutions having both spatial and temporal correlations~\cite{Forrester2009}. Once trained, they can be used to replace high-fidelity simulations in many query applications without incurring additional costs. ROMs require high upfront offline training costs to allow rapid evaluations with minimal online costs. 
 
The main objective of ROM is to find a low-dimensional representation, called a latent space or subspace, that best describes the original high-dimensional solution space.  This often takes the form of a linear subspace such that the transformation from the high- to the low-dimensional field is given by a projection map. 

Non-intrusive ROM methods are frequently used in engineering applications due to their versatility and ease of implementation~\cite{Xiao2015Non-intrusiveEquations}. This category of methods treats the original simulation model as a black-box function and does not require access to the underlying simulation source code. Proper orthogonal decomposition (POD) is a simple and efficient dimensionality reduction method that is often combined with a regression model to form a non-intrusive ROM~\cite{lu2019review,Fossati2015EvaluationMethodology, decker2020dimensionality,rajaram2020randomized,Behere2021ReducedEstimation}. This POD-regressor combination can be trained to predict new solutions in latent space. The linearity of the POD technique can then be used to map the predicted solution from latent space to output dimension space. Recently, deep learning (DL)-based ROM techniques have emerged as popular tools due to their prowess in capturing complex nonlinear relationships between inputs and outputs~\cite{Chen2023TowardsAerofoils,Kashefi2021AGeometries,Deng2023PredictionStrategies}. However, it's important to note that DL-based ROMs require a substantial amount of training data to effectively learn these relationships. A recent study by~\citet{mufti2024shock} demonstrated that linear POD-based ROMs hold an advantage over DL-based ROMs when the number of training samples is limited.

\subsection{ROMs with High-Dimensional Input}

To effectively capture the variability within the design space, engineering models that are represented by ROMs and scalar-based data-fit surrogate models are often parameterized by a large number of design variables.
As a result, they tend to suffer from the \emph{curse of dimensionality}~\cite{Koch1999}, a common expression used to illustrate the challenges associated with high-dimensional problems. This results in a rapid increase in the volume of the design space with increasing design variables, requiring a disproportionately large number of samples for a dense coverage of the design space. High-dimensional input spaces also make it harder to reason on distances when defining input-output relationships, a phenomenon sometimes referred to as the \emph{concentration of distances}~\cite{Kumari2017}.
%This severely affects the accuracy of these models. Thus, the creation of such models motivates the application of both reduced order modeling and parameter space reduction techniques, i.e., both input and output dimensionality reduction 
%ROMs and surrogate models suffer from the curse of dimensionality when the number of input dimensions is large.
Input dimensionality reduction (DR) techniques can be used to effectively alleviate this curse of dimensionality. Unsupervised dimension reduction techniques such as principal component analysis (PCA)~\cite{Brunton2019AnnualMechanics}, diffusion maps~\cite{dietrich2018fast}, etc., work decently when the input variables are correlated. However, in most ROMs or surrogate modeling applications, the input space is usually sampled uniformly with a quasi-random design of experiment (DOE). For such problems, no structure in the inputs alone can be exploited, and a supervised DR technique that leverages instead the explicit input-output relationship is best suited. Sensitivity analysis~\cite{saltelli2008global}, automatic relevance determination~\cite{neal1998assessing}, and partial least squares~\cite{Bouhlel2016} are some of the supervised DR techniques that have been used to carry out input space dimension reduction with limited success.

Active subspace methodology (ASM) is a tool that has recently gained success due to its ability to effectively reduce input dimensionality. The methodology identifies a low dimensional linear subspace, referred to as the active subspace, that best describes the variability in the output due to changes in input variables \cite{Constantine2014,Berguin2015DimensionalityGradient,Berguin2015MethodAnalyses,Jiang2020AQuantification}. The classic ASM methodology applies eigenvalue decomposition on gradient information to find this subspace. It has been successfully applied to solve aerodynamic shape optimization~\cite{Lukaczyk2014,Li2019Surrogate-basedMethod}, hull design \cite{tezzele2019shape}, cavity design to stabilize supersonic flames~\cite{wei2023exploiting}, hyper-sonic engine uncertainty quantification~\cite{Constantine2015}, materials design~\cite{Khatamsaz2021AdaptiveDesign} and sensitivity analysis for lithium-ion battery~\cite{Constantine2017Time-dependentModel}. A multi-fidelity version of ASM can be used to reduce the cost of input dimension reduction using active subspace~\cite{Lam2020MultifidelitySubspaces,romor2023multi, Mufti2022AInputs}. However, obtaining gradient information through intrusive methods from a simulation is not always straightforward, especially for closed-source tools that do not readily provide this capability. This limitation can restrict the application of the active subspace methodology. To address this challenge, gradient-free methods have been proposed, which identify the active subspace using only direct function evaluations. Some of these gradient-free methods use probabilistic techniques, such as Gaussian process regressor (GPR) with built-in dimensionality reduction \cite{Tripathy2016GaussianPropagation,Tsilifis2021BayesianProcesses,gautier2022fully}. The active subspace projection matrix is treated as an additional hyper-parameter to be estimated from the given data. 

\begin{figure}[ht]
    \centering
    \includegraphics[width=1\textwidth]{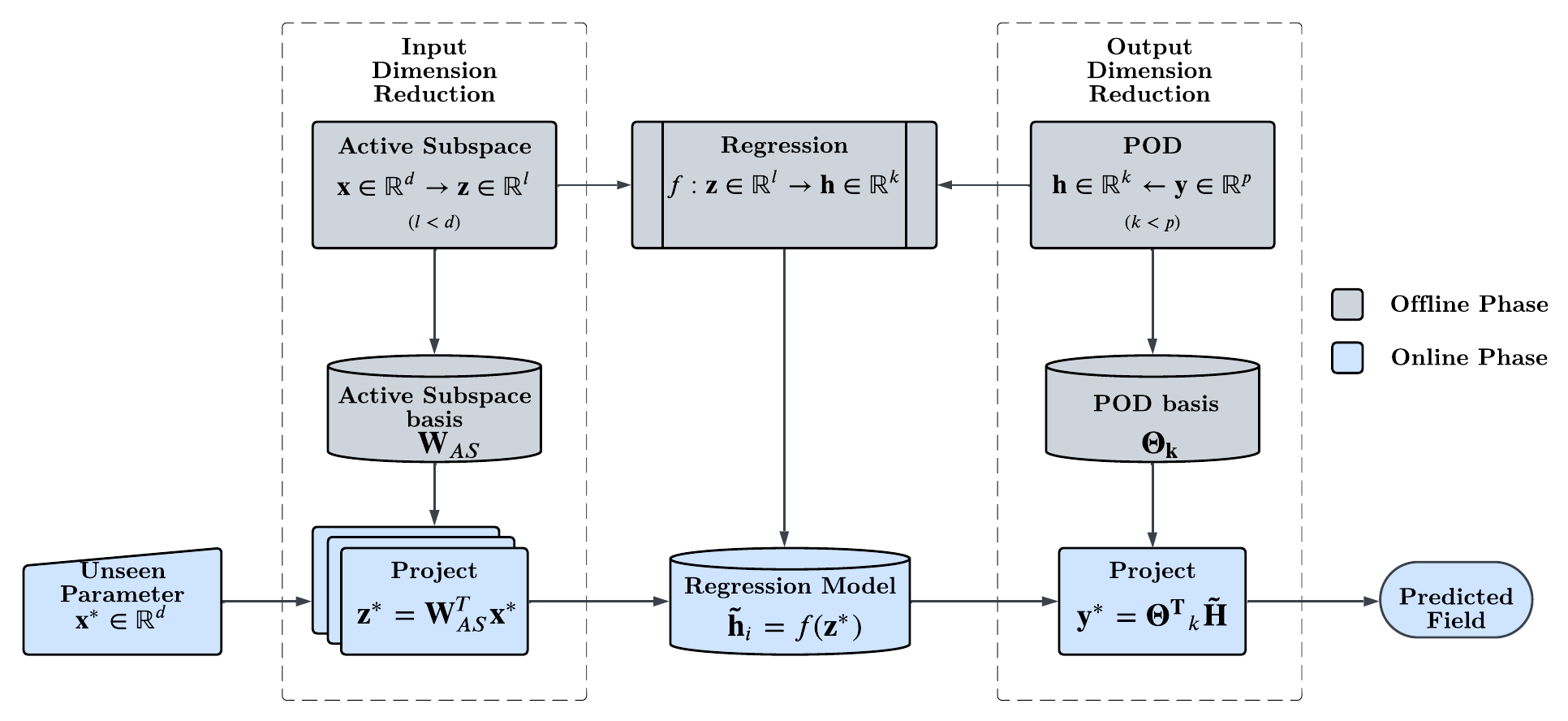}
    \caption{A graphical representation of the PCAS method to develop ROMs for high-dimensional input space.}
    \label{fig:PCAS}
\end{figure}

The input DR methods discussed previously have been effectively used to create scalar-based data-fit surrogate models. However, constructing ROMs with high-dimensional input space poses additional challenges: (1) we need to find an active subspace in which the low-dimensional representation of the output field varies the most, and (2) a separate active subspace has to be computed for each of the projected coordinates in the low-dimensional output space. The PCAS technique has been successfully used to create ROMs for engineering problems with high-dimensional input space. Fig.~\ref{fig:PCAS} gives a graphical representation of steps involved in the application of the PCAS model to construct ROMs.  The methodology first carries out output space reduction using POD. A  linear regression model is then created linking POD modes and the input space. Gradient information is found using this regression model. The ASM is then used to find the active subspace in the input domain. PCAS has been used in additive manufacturing~\cite{Vohra2020FastManufacturing}, aircraft fuselage panel analysis~\cite{guo2023investigation}, beam structural analysis~\cite{demo2019non} and ship hull design~\cite{tezzele2019shape}. A modified version of this method called PCAS-AK combines PCAS with active Kriging to adaptively select training samples to improve computational efficiency~\cite{Ji2022HighTechnique}.  Recently, a non-intrusive POD-based ROM which uses a GPR with an in-built input space reduction~\cite{Rajaram2020}, and derivative-informed projected neural networks (DIPNets)~\cite{OLeary-Roseberry2022Derivative-informedPDEs} have also been used to develop a prediction model for problems with high-dimensional input and output space.

\subsection{Multi-Fidelity ROM Techniques}
A major drawback of existing ROM methods for high dimensional input space is that they require substantial training data to achieve reasonable accuracy~\cite{Ji2022HighTechnique}. For most engineering problems, generating a large quantity of high-fidelity simulation data can become cost-prohibitive.
Multi-fidelity methods alleviate the high training cost of creating such ROMs by leveraging cheaper to evaluate low-fidelity data \cite{Forrester2007Multi-fidelityModelling,peherstorfer2018survey}. Multi-fidelity ROM techniques use dense low-fidelity data to improve their prediction accuracy with a modest increase in computational budget. Most multi-fidelity ROM techniques necessitate consistent representations between field solutions of various fidelities, meaning they require a similar discretization of the output field. Such methods use different numerical methods for different fidelity levels \cite{Bertram2018TowardsModeling}. Manifold alignment \cite{wang2011manifold} was recently introduced as a technique that can be used to align underlying latent spaces of various fidelity levels having inconsistent representation  \cite{perron2020multi}. This technique was combined with a POD-regressor combination to construct a multi-fidelity manifold aligned ROM (MA-ROM) and was shown to offer a computational cost and accuracy advantage over its single-fidelity counterpart in predicting pressure and stress fields \cite{Perron2021Multi-fidelityAlignment,Perron2022ManifoldAnalysis}. Most of the multi-fidelity ROM techniques are limited by the number of input variables that can be used. They have been developed for problems with 1 to 10 design variables \cite{Decker2023ManifoldModeling, Perron2022ManifoldAnalysis}. The need for developing multi-fidelity ROMs for high-dimensional design spaces has been recognized by several authors \cite{li2022machine,benner2015survey}.

\subsection{Contributions of Current Work}
This study proposes a method that combines the machine learning concept of manifold alignment with dimension reduction via POD and ASM to address the limitations of existing methods. This method, applied in a multi-fidelity setting, creates a non-intrusive ROM for problems with high dimensional input spaces.  The main contributions of this study are: (1) it develops a multi-fidelity, computationally efficient, and simple approach for constructing ROMs for high dimensional input space, and (2) it uses a multi-fidelity model-based active subspace approach to carry out input dimensional reduction. The sensitivity of the proposed methodology is assessed by: (a) varying the number of training samples,  (b) changing the ratio of high- and low-fidelity data in training samples, and (c) varying the total number of input dimensions. The proposed method is also benchmarked against existing ROM techniques.

The rest of the paper is organized as follows. In Sec.~\ref{sct: methodology}, a detailed explanation of the proposed approach is given followed by the formalization of the method. Sec.~\ref{sct: application} introduces the performance metrics and test cases that will be used to evaluate the performance of the proposed method. Sec.~\ref{sct: results}  discusses the results of the experiments and establishes the computational efficiency of the method by comparing its performance with the single-fidelity PCAS and MA-ROM methods.
\section{Methodology}\label{sct: methodology}
This section provides details of the numerical method developed in this study. An overview of the Procrustes manifold alignment technique, which is used for output dimension reduction is given. Then an explanation of a multi-fidelity model-based active subspace technique used to reduce input dimensionality is presented. Finally, the two methods are integrated to create a multi-fidelity ROM for high-dimensional input space.

\subsection{Output Dimension Reduction using Procrustes Manifold Alignment}
Manifold alignment (MA) is a machine learning technique that is used to find a unifying representation of multiple datasets that share a common low-dimensional latent space. Manifold alignment works on the principle of transfer learning. The technique has been used to solve problems in fields of bio-informatics~\cite{wang2009general}, medical imaging~\cite{guerrero2014manifold},  computer vision~\cite{bousmalis2016domain}, etc. Procrustes analysis is an approach used for manifold alignment. The analysis aligns the latent space of multiple datasets which can be found using a dimensionality reduction technique~\cite{wang2008manifold}. In the proposed method, Procrustes manifold alignment is combined with POD to carry out dimension reduction and alignment of different fidelity output fields as demonstrated by~\citet{Perron2022ManifoldAnalysis}.

\subsubsection{Mathematical Formulation}
Let us consider two datasets $\mathbf{Y} \in \mathbb{R}^{p \times m_{1}}$ and $\mathbf{Z} \in \mathbb{R}^{q \times m_{2}}$, with $m_{2} > m_{1}$. Here,  $m_1$, and $m_2$ represent the number of samples, and $p$ and $q$ represent the dimensionality of datasets.  The datasets $\mathbf{Y}$ and $\mathbf{Z}$ represent simulation results from a high- and low-fidelity analysis respectively with the columns of the matrices holding field solutions.
It is assumed that these datasets are related in some way, e.g., they share similar governing equations but have different field discretization. We can consider both $\mathbf{Y}$ and $\mathbf{Z}$ to be simulation results for an input parameter space $\mathcal{X} \in \mathbb{R}^{d}$, where \textit{d} is the dimension of the input space. Additionally, we can assume that the first $m_1$ solutions in both datasets were generated using the same input parameter values. In other words, these $m_1$ data points are considered \emph{linked} while the remainder is assumed to be \emph{unlinked}.

The first step of Procrustes manifold alignment is to find the low dimensional latent space for both the high- and low-fidelity datasets individually. In the current case, this is done using the conventional POD method and produces the following results
\begin{align}\label{eq:x=Wz}
\mathbf{H} &= \boldsymbol{\Theta}_k^T \mathbf{Y}\\
\mathbf{S} &= \boldsymbol{\Xi}_k^T \mathbf{Z}
\end{align}
where $\mathbf{H} \in \mathbb{R}^{k_{1} \times m_{1}}$ and $\mathbf{S}  \in \mathbb{R}^{k_{2} \times m_{2}}$ contains latent variables. The matrices $\boldsymbol{\Theta}_k \in \mathbb{R}^{p \times k_{1}}$ and $\boldsymbol{\Xi}_k \in \mathbb{R}^{q \times k_{2}}$ represent the orthonormal bases corresponding to the largest eigenvalues of $\mathbf{Y}$ and $\mathbf{Z}$ respectively, where $(k_{1},k_{2}) \ll (p,q)$. The dimensionality ($k_{1},k_{2}$) of both high- and low-fidelity latent space can be found using relative information criterion (RIC) \cite{pinnau2008model}, defined as :
\begin{equation}\label{eq:RIC}
\text{RIC} = \frac{\sum_{i=1}^{k} \lambda_i}{\sum_{j=1}^{p|q} \lambda_j}
\end{equation}
In this formulation, we will assume $k_{1} = k_{2} =  k$, but in some cases, due to a large number of low-fidelity samples, $k_{2}$ can be larger than $k_{1}$.
For such cases, the Procrustes analysis is still applicable, but the alignment problem turns into a projection Procrustes problem \cite{gower2010procrustes}.

We can partition the low-fidelity latent variables into linked and unlinked datasets, i.e. $\mathbf{S} = [\mathbf{S}_L, \mathbf{S}_U]$ where $\mathbf{S}_L \in \mathbb{R}^{k \times m_{1}}$ and $\mathbf{S}_U \in \mathbb{R}^{k \times (m_{2} - m_{1})}$. Once both datasets have been projected to a reduced dimensional latent space, the next step is to find an affine transformation (i.e., translation, rotation, reflection, scaling) to align latent variables sets having the same input parameter values, i.e., align $\mathbf{S_{L}}$ with $\mathbf{H}$. This can be achieved with the following steps:

\begin{enumerate}
    \item Translate $\mathbf{S_{L}}$ and $\mathbf{H}$ such that their respective mean is zero, i.e., centroids of the datasets should be at the origin.\\
    
    \item Compute the SVD of $\mathbf{S_{L}}\mathbf{H}$ as
     \begin{equation}
    \mathbf{U} \boldsymbol{\Sigma} \mathbf{V}^T = \mathbf{S}_L \mathbf{H}^T
    \end{equation}
    Where,
    
    $\mathbf{\Sigma} \in \mathbb{R}^{k \times k}$ is a diagonal matrix containing the singular values.
    
    $\mathbf{U} \in \mathbb{R}^{k \times k}$ and $\mathbf{V} \in \mathbb{R}^{k \times k}$ are square matrices containing the left and right singular vectors respectively.\\
    
    \item Using the left and right singular vectors we can find the optimal transformation matrix as:
    \begin{align}
    \mathbf{P} &= \mathbf{V}\mathbf{U}^T
    \end{align}
    
    \item The scaling factor can be calculated using the eigenvalue matrix
    \begin{align}
    t = \frac{\text{trace}(\boldsymbol{\Sigma})}{\text{trace}\left(\mathbf{S}_L \mathbf{S}_L^T\right)} 
    \end{align}
    
    \item As a final step, the low-fidelity latent variables that are aligned with high-fidelity variables can be computed as:
    \begin{equation}\label{eq:MA-mapping}
    \mathbf{G} = t \mathbf{P} \mathbf{S}
    \end{equation}
\end{enumerate}
The main steps of the Procrustes manifold alignment algorithm are outlined in Algorithm \ref{alg:Procrustes}. At the end of the output dimension reduction phase, we have two datasets $\mathbf{H}$ and $\mathbf{G}$ containing latent variables of the high- and low-fidelity data respectively. The dimensionality of both these data sets is the same, i.e.,  $\mathbf{H} \in \mathbb{R}^{k \times m_{1}}$ and $\mathbf{G} \in \mathbb{R}^{k \times m_{2}}$, and they lie on the same manifold. We only need to store $\mathbf{H}$, $\mathbf{G}$, and $\boldsymbol{\Theta}_k$, at the end of this phase. $\mathbf{H}$ and $\mathbf{G}$ will be used to construct an active subspace for each latent variable, and $\boldsymbol{\Theta}_k$ will be used to reconstruct the output field of interest.

\begin{algorithm}
    \caption{Procrustes Manifold Alignment}
    \label{alg:Procrustes}
    \textbf{Input:} High-fidelity data $\mathbf{Y} \in \mathbb{R}^{p \times m_1}$, low-fidelity data $\mathbf{Z} \in \mathbb{R}^{q \times m_2}$ \\
    \textbf{Output:}  Aligned latent variables $\mathbf{H}$, $\mathbf{G}$, and high-fidelity POD modes $\boldsymbol{\Theta}_k$ 
    
    \begin{algorithmic}[1]
        \STATE Find orthonormal bases $\boldsymbol{\Theta}_k \in \mathbb{R}^{p \times k}$ and $\boldsymbol{\Xi}_k \in \mathbb{R}^{q \times k}$ using the largest eigenvalues of $\mathbf{Y}$ and $\mathbf{Z}$, respectively
        \STATE Compute low-dimensional latent variables $\mathbf{H} = \boldsymbol{\Theta}_k^T \mathbf{Y}$ and $\mathbf{S} = \boldsymbol{\Xi}_k^T \mathbf{Z}$
        \STATE Partition $\mathbf{S}$ into $\mathbf{S} = [\mathbf{S}_L, \mathbf{S}_U]$, where $\mathbf{S}_L \in \mathbb{R}^{k \times m_1}$ and $\mathbf{S}_U \in \mathbb{R}^{k \times (m_2 - m_1)}$
        \STATE Translate $\mathbf{S}_L$ and $\mathbf{H}$ such that their respective mean is zero
        \STATE Compute the SVD of $\mathbf{S}_L\mathbf{H}^T$: $\mathbf{U} \boldsymbol{\Sigma} \mathbf{V}^T = \mathbf{S}_L \mathbf{H}^T$
        \STATE Compute optimal transformation matrix: $\mathbf{P} = \mathbf{V}\mathbf{U}^T$
        \STATE Compute scaling factor: $t = \frac{\text{trace}(\boldsymbol{\Sigma})}{\text{trace}\left(\mathbf{S}_L \mathbf{S}_L^T\right)}$
        \STATE Compute low-fidelity latent variables aligned with high-fidelity variables: $\mathbf{G} = t\mathbf{P}\mathbf{S}$
        \STATE \textbf{return} $\mathbf{H}$, $\mathbf{G}$,$\boldsymbol{\Theta}_k$
    \end{algorithmic}
\end{algorithm}

\subsubsection{Multi-Fidelity Reduced Order Model}
We can create a multi-fidelity reduced-order model that uses both high- and low-fidelity latent variables  $\mathbf{H}$ and $\mathbf{G}$, to predict the new high-fidelity field solutions. This manifold-aligned ROM (MA-ROM) can be constructed using any multi-fidelity regression technique \cite{Choi2009Two-levelJets, Han2012AlternativeModeling,Han2012} and will be computationally efficient as compared to the corresponding single-fidelity ROM. However, this ROM would have an input domain $\mathcal{X}$, which is still high-dimensional and hence would suffer from the \textit{curse of dimensionality}.

\subsection{Input Dimension Reduction using Multi-Fidelity Model-Based Active Subspace}
To reduce input dimensionality, we want to find a low dimensional subspace that effectively captures the variability in output latent space due to change in $\mathcal{X}$. This active subspace is spanned by the dominant eigenvectors of the estimator for the covariance (second moment) matrix. To find the multi-fidelity approximation of this covariance matrix we need to find gradient information of latent variables in $\mathbf{H}$ and $\mathbf{G}$ with $\mathcal{X}$. This gradient information can be obtained using adjoint methods, automatic differentiation, or finite differencing. However, implementing adjoint formulation or automatic differentiation becomes unfeasible for proprietary or closed-source simulation tools unless these methods are already in place. On the other hand, finite difference approximations, although more broadly applicable, tend to scale poorly as the number of input variables increases. The application of these methods is also computationally expensive as we need to find $2k$ gradient computations corresponding to each latent variable in $\mathbf{H}$ and $\mathbf{G}$. Instead, we will use a multi-fidelity model-based active subspace method, a simple but effective technique to find the active subspace. This technique has previously been used to construct scalar-based surrogate models for high-dimensional inputs and is modified here for applications in ROMs~\cite{mufti2023design}.

\subsubsection{Mathematical Formulation}
It is important to remember that we are interested in predicting high-fidelity field solutions by estimating their latent space coordinates using multi-fidelity information. We know that each row in the latent variable matrices $\mathbf{H}$ and $\mathbf{G}$ corresponds to variability in a feature due to a change in inputs in $\mathcal{X}$. We can consider each feature $\mathbf{h}_i$ and $\mathbf{g}_i$  as a scalar-valued function of the set of inputs. An active subspace in the context of this problem is a low-dimensional subspace in which most of the variations in $\mathbf{h}_i$ and $\mathbf{g}_i$ due to variability in $\mathcal{X}$ are concentrated. 

We can predict our quantity of interest using a multi-fidelity surrogate modeling framework that uses additive correction \cite{Choi2009Two-levelJets}.  
\begin{equation}\label{eq:additive_correction}
    \Tilde{h}_i(\mathbf{x}) = g_i(\mathbf{x}) + \delta(\mathbf{x})
\end{equation} 
where $\delta(\mathbf{x})$ is the correction term that can be determined by finding the difference between high- and low-fidelity latent variables at $m_1$ linked data points. We can find $\Tilde{h}_i(\mathbf{x})$ at unknown points by replacing $g_i(\mathbf{x})$ and $\delta(\mathbf{x})$ with a surrogate model i.e. $\Tilde{g}_i(\mathbf{x})$ and $\Tilde{\delta}(\mathbf{x})$. The availability of high-fidelity points is limited due to computational cost considerations. Hence, we will model $\Tilde{\delta}(\mathbf{x})$ using a linear model

\begin{equation}\label{eq:linear_model}
   \Tilde{\delta}(\mathbf{x}) = a + \mathbf{b}^T\mathbf{x}
\end{equation}
where $a$ and $\mathbf{b}$ can be found by solving the linear least squares problem
\begin{equation}
    \underset{a\in \mathbb{R}, \mathbf{b} \in \mathbb{R}^d}{\text{min}} \sum_{j=1}^{m_{1}}\left[(h_{ij} - g_{ij}) - (a + \mathbf{b}^T \mathbf{x}_j)\right]^2
\end{equation}
We would like to highlight at this point that the multi-fidelity model $\Tilde{h}$ from Eq.~\eqref{eq:additive_correction}, will be used to find the gradient information needed to develop the covariance matrix to find an active subspace. We will not use $\Tilde{h}$ to directly predict the latent coordinates of our output field of interest. We just require $\Tilde{h}$ to be accurate enough to find variations in output latent space coordinates with changes in inputs. 

We can generate a relatively large number of low-fidelity data points. Depending on the cost of obtaining low-fidelity samples, the low-fidelity solution in Eq.~\eqref{eq:additive_correction} can be replaced with a linear model, a higher-order polynomial or a radial basis function (RBF) surrogate model. We will use an RBF surrogate with a non-parametric basis function in this study to represent the low-fidelity data. Non-parametric basis function does not contain hyper-parameters. Hence, the process of finding these hyper-parameters which scale poorly with the increasing number of input variables can be avoided. The mathematical expression for an RBF surrogate model is given as

\begin{equation}
    \Tilde{g}(\mathbf{x}) = \sum_{i=1}^{m_{2}} w_{i} \phi( \lVert \mathbf{x} - \mathbf{x}_{i} \rVert) + \epsilon(\mathbf{x})
\end{equation}
where $w_i$ are the weights of the surrogate model, and $\phi$ represents the basis function. The polynomial trend is represented by $\epsilon(\mathbf{x})$. 
We can now find the multi-fidelity gradient information 
\begin{equation}\label{eq:gradient_approx}
    \nabla \Tilde{h}(\mathbf{x}) =  \nabla \Tilde{g}(\mathbf{x}) + \nabla \Tilde{\delta}(\mathbf{x})
    = \mathbf{b} + (\epsilon'(\mathbf{x}) +\mathbf{\Phi'}\mathbf{w})
\end{equation} 
Here, $\epsilon'(\mathbf{x}$) is the gradient of the polynomial trend, and $\mathbf{\Phi'}$ represents the gradient of the kernel matrix. For a non-parametric cubic kernel, $\phi(\mathbf{x},\mathbf{x}_i) = \lVert \mathbf{x} - \mathbf{x}_{i} \rVert^3$, gradient of kernel matrix can be found out using
\begin{equation}
    \mathbf{\Phi'} = 3 \lVert \mathbf{x} - \mathbf{x}_{i} \rVert (\mathbf{x} - \mathbf{x}_{i})
\end{equation}
The covariance matrix can then be found using the outer product of the gradient with itself.
\begin{equation}\label{eq:matrix_H_integral}
    \mathbf{C}_{MF}=\int \nabla \Tilde{h}(\mathbf{x}) \nabla \Tilde{h}(\mathbf{x})^{T}\rho(\mathbf{x})\,dx
\end{equation} 
\begin{equation}\label{eq:rbf_integral}
    \mathbf{{C}}_{MF} = \int \left(\mathbf{b} + (\epsilon'(\mathbf{x}) +\mathbf{\Phi'}\mathbf{w}\right))\left(\mathbf{b} + (\epsilon'(\mathbf{x}) +\mathbf{\Phi'}\mathbf{w}\right))^{T}\rho(\mathbf{x})\,dx 
\end{equation}
where $\rho(\mathbf{x})$ is the probability density on $\mathcal{X}$. We can solve the integration in Eq.~\eqref{eq:rbf_integral} numerically to obtain $\mathbf{{C}}_{MF}$. The covariance estimator is multi-fidelity in nature as the gradient information used has been approximated using a multi-fidelity framework.  Since matrix $\mathbf{{C}}_{MF}$ is symmetric, it can be diagonalized as 
\begin{equation}
    \mathbf{{C}}_{MF}=\mathbf{{W}{\Lambda}} {\mathbf{W}}^T
\end{equation}
In the above equation, $\mathbf{{\Lambda}}=diag(\lambda_1,\lambda_1,\dots,\lambda_d)$ is a diagonal matrix whose elements are the eigenvalues of $\mathbf{{C}_{MF}}$ organized in descending order, i.e.,  $\lambda_1 \geq \cdots \geq \lambda_d \geq 0$.
Also, 
$\mathbf{{W}}=[\mathbf{w}_1,\dots,\mathbf{w}_d]$ is a orthogonal matrix where the $i$-th column $\mathbf{w}_i$ is the eigenvector corresponding to $\lambda_i$.
A higher eigenvalue here means a greater variation in response gradient, we can examine the eigenvalue decay or use a pre-defined criterion to separate the largest $l$ eigenvalues with their corresponding eigenvectors such that
\begin{equation}\label{eq:eigen_sep}
    \mathbf{
    {\Lambda}}=
    \begin{bmatrix}
        \boldsymbol{{\Gamma}}_{AS} & 0 \\
        0 & \boldsymbol{{\Gamma}}_\perp\\
    \end{bmatrix},\;   
    \mathbf{{W}}=
    \begin{bmatrix}
        \mathbf{{W}}_{AS} & \mathbf{W}_\perp \\
    \end{bmatrix}
\end{equation}    
The selected $l$ eigenvectors $\mathbf{{W}}_{AS} =\left\{\mathbf{w}_i\, |\, i=1,\dots,l\right\}$ form the active subspace while the remaining $d-l$ eigenvectors are defined as the inactive subspace.

The multi-fidelity model-based active subspace approach is expected to be computationally advantageous as~(1) it effectively reduces the input dimension mitigating the effects of the \textit{curse of dimensionality}, and~(2) it uses cheap-to-evaluate surrogate models to predict gradient information. The methodology is expected to work best in cases where high-fidelity simulation output is expensive to evaluate and low-fidelity simulation is a good approximation of it. The approach also assumes the existence of an active subspace, for most engineering problems this is a valid assumption as they generally exhibit low-dimensional structure in their input-output relationships. The main steps of the multi-fidelity model-based active subspace algorithm are outlined in the Algorithm \ref{alg:MBAS}.

\begin{algorithm}
    \caption{Multi-Fidelity Model-Based Active Subspace}
    \label{alg:MBAS}
    \textbf{Input:} $m_1$ samples of $h(\mathbf{x})$, $m_2$ samples of $g(\mathbf{x})$ \\
    \textbf{Output:} Set of eigenvectors $\mathbf{{W}}_{AS}$ forming an active subspace
    \begin{algorithmic}[1]
        \STATE Represent high-fidelity quantity of interest as  $\Tilde{h}(\mathbf{x}) = \Tilde{g}(\mathbf{x}) + \Tilde{\delta}(\mathbf{x})$
        \STATE Compute correction term $\Tilde{\delta}(\mathbf{x}) = a + \mathbf{b}^T\mathbf{x}$, where $a$, $\mathbf{b}$ are minimizers of:$$\underset{a\in \mathbb{R}, \mathbf{b} \in \mathbb{R}^d}{\text{min}} \sum_{j=1}^{m_{1}}\left[(h_{j} - g_{j}) - (a + \mathbf{b}^T \mathbf{x}_j)\right]^2$$
        \STATE Approximate $\Tilde{g}(\mathbf{x})$ as:$$\Tilde{g}(\mathbf{x}) = \sum_{i=1}^{m_2} w_i \phi(||\mathbf{x}-\mathbf{x}_i||) + \epsilon(\mathbf{x})$$
        \STATE Compute gradient $\nabla \Tilde{h}(\mathbf{x})= \mathbf{b} + (\epsilon'(\mathbf{x}) +\mathbf{\Phi'}\mathbf{w})$
        \STATE Compute the eigenpairs of covariance matrix $\mathbf{C}_{MF}$ using:
        $$\mathbf{C}_{MF}=\int \nabla \Tilde{h}(\mathbf{x}) \nabla \Tilde{h}(\mathbf{x})^{T}\rho(\mathbf{x})\,dx = \mathbf{{W}{\Lambda}} {\mathbf{W}}^T$$
        \STATE Select $l$ eigenvectors $\mathbf{{W}}_{AS} =\left\{\mathbf{w}_i\, |\, i=1,\dots,l\right\}$ using the eigenvalue spectrum.
        \STATE \textbf{return} $\mathbf{{W}}_{AS}$
    \end{algorithmic}
\end{algorithm}

\subsection{Multi-Fidelity Reduced Order Model in Active Subspace}
After the application of multi-fidelity model-based active subspace, the dimension reduction in input space is $\mathbb{R}^{d} \mapsto \mathbb{R}^{l}$, where $l < d$. We can now construct a multi-fidelity regression model mapping low-dimensional input space to high and low-fidelity latent variables, i.e. $\mathbf{h}_i$ and $\mathbf{g}_i$.  In this study, we will use Hierarchical Kriging (HK) \cite{Han2012HierarchicalModeling}, a relatively simple and efficient multi-fidelity regression technique. The formulation for HK is given by
\begin{equation}\label{eq:regression}
    \tilde{h}_i(\mathbf{\xi}) = \alpha \tilde{g}_i(\mathbf{\xi}) + \mathbf{\omega^{T}} \mathbf{r(\mathbf{\xi})}
    \end{equation}
where, 
$\mathbf{\xi} = \mathbf{{W}}_{AS_{i}} \mathbf{x}$ is the reduced dimension input space,
$\tilde{h}_i$ and $\tilde{g}_i$ are Kriging predictors of high and low-fidelity latent variables respectively,
$\alpha$ is a scaling factor, $\mathbf{\omega}$ is a weight vector, and $\mathbf{r(\mathbf{\xi})}$ is a kernel function.

A separate multi-fidelity active subspace is computed for each latent variable in $\mathbf{H}$ and a corresponding surrogate model fit $\tilde{h}_i$ is carried out i.e. a total of $k$ different multi-fidelity active subspaces and multi-fidelity regression models are constructed. This multi-fidelity regression model will suffer less from the curse of dimensionality as it has been built over a reduced dimensional input space. New high-fidelity field solutions can now be predicted by estimating their latent variables using Eq.~\eqref{eq:regression} and mapping them back to output space using high-fidelity POD modes $\boldsymbol{\Theta}_k$.
\begin{equation}
    \mathbf{\tilde{y}}=\sum_{i=1}^{k}\boldsymbol{\theta}_i\tilde{h}_i(\mathbf{\xi})
\end{equation}

\subsection{Proposed Multi-Fidelity ROM Method for High-Dimensional Input Space}

\begin{figure}[ht]
    \centering
    \includegraphics[width=1\textwidth]{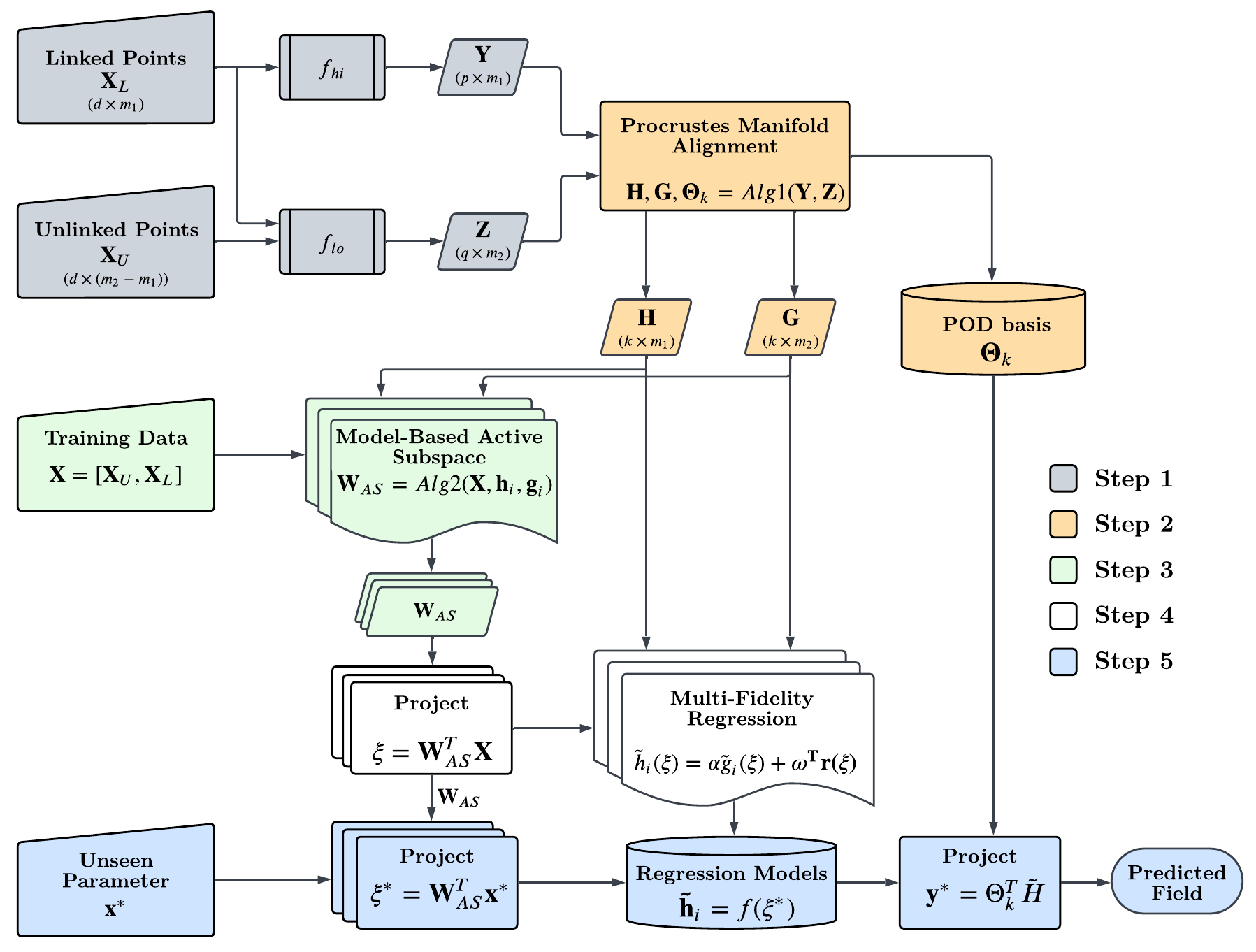}
    \caption{A graphical representation of the MF-PCAS method to develop ROMs for high-dimensional design space.}
    \label{fig:MF-PCAS}
\end{figure}

The main contribution of this work is the formulation of a computationally efficient, multi-fidelity, non-intrusive ROM for problems having high-dimensional design space. The methodology uses Procrustes manifold alignment to carry out output dimension reduction. The \textit{curse of dimensionality} is then alleviated by using a multi-fidelity model-based active subspace approach. The proposed methodology is henceforth referred to as MF-PCAS. The overall procedure for constructing an MF-PCAS model is illustrated in Algorithm~\ref{alg:MF-PCAS} and in Fig.~\ref{fig:MF-PCAS}, and is summarized below: 
% Some other Possible options for names MFPCAS, Manifold aligned active subspaces (MA-AS). Procrustes-Aligned Multi-Fidelity Active Subspace" (PAMAS). Multi-Fidelity Procrustes Active Subspace Model (MF-PRASM).
\\
\begin{enumerate} [itemsep=1.5\baselineskip]
    \item \textbf{Generate Data:} Sample $m_1$ designs from high-dimensional input space $\mathcal{X}$. Run high- and low-fidelity models to generate matrices $\mathbf{Y}$ and $\mathbf{Z}_L$ containing field solutions in their columns. Sample an additional $m_2 - m_1$ design from $\mathcal{X}$, and run only a low-fidelity model to generate matrix $\mathbf{Z}_U$.
    
    \item \textbf{Reduce Output Dimensionality Using Procrustes Manifold Alignment:} Apply Procrustes manifold alignment to align the low-dimensional latent spaces of matrices $\mathbf{Y}$ and $\mathbf{Z} =[\mathbf{Z}_L, \mathbf{Z}_U]$. This can be achieved with the following sub-steps:
    \begin{enumerate} [label=(\alph*)]
        \item Apply POD to obtain the low-dimensional latent variables $\mathbf{H}$ and $\mathbf{S} = [\mathbf{S}_L,\mathbf{S}_U]$, as well as the orthonormal bases $\boldsymbol{\Theta}_k$ and $\boldsymbol{\Xi}_k$ for high- and low-fidelity fields respectively.
        \item Find an optimal translation vector, transformation matrix, and scaling factor by performing the Procrustes analysis of $\mathbf{H}$ and $\mathbf{S}_L$.
        \item Get the low-fidelity latent variables $\mathbf{G}$ by using the calculated optimal transformation, ensuring that they have the same number of dimensions and are located on the same manifold as $\mathbf{H}$.
    \end{enumerate}

    \item \textbf{Reduce Input Dimensionality Using Multi-Fidelity Model-Based Active Subspace:} Find a separate active subspace for each latent variable in $\mathbf{H}$ using multi-fidelity model-based heuristics. The process is described in the following sub-steps:
    \begin{enumerate} [label=(\alph*)]
        \item Develop a multi-fidelity surrogate modeling framework to predict the high-fidelity latent variable $\Tilde{h}_i(\mathbf{x})$, using the low-fidelity latent variable $\Tilde{g}_i(\mathbf{x})$ and a discrepancy term $\Tilde{\delta}(\mathbf{x})$.
        \item Find the covariance matrix $\mathbf{{C}}_{MF}$ using the outer product of gradient $\nabla\Tilde{h}_i(\mathbf{x})$ with itself, and apply eigenvalue decomposition to it.
        \item Analyze the eigenvalue spectrum to find the first $l$ eigenvectors that define the active subspace $\mathbf{{W}}_{AS}$.
        \item Repeat the process $k$ times to find a different active subspace for each latent variable in a $k$-dimensional latent variable matrix $\mathbf{H}$.
    \end{enumerate}

    \item \textbf{Train Regression Model:} Train a set of $k$ different hierarchical Kriging models $\mathbf{\tilde{h}}(\mathbf{\xi}) = [\tilde{h}_1(\mathbf{\xi}_1), \dots, \tilde{h}_k(\mathbf{\xi}_k)]$. These models are built over reduced dimensional input space $\mathbf{\xi}_i = \mathbf{{W}}_{AS_i} \mathbf{x}$, and combine information from both $\mathbf{H}$ and $\mathbf{G}$.

    \item \textbf{Predict Output Field of Interest:} For an out-of-sample design point $\mathbf{x}^*$,  we can predict output field using following steps:
    \begin{enumerate}[label=(\alph*)]
        \item Predict the corresponding latent variable using $\mathbf{\tilde{h}^*}(\mathbf{\xi^*}) = [\tilde{h}^*_1(\mathbf{\xi^*}_1), \dots, \tilde{h}^*_k(\mathbf{\xi}_k)]$, where $\mathbf{\xi}^*_i = \mathbf{{W}}_{AS_i} \mathbf{x}^*$.
        \item Reconstruct the output field using $\boldsymbol{\Theta}_k$ and $\mathbf{\tilde{h}^*}$.
    \end{enumerate}
\end{enumerate}

\begin{algorithm}
    \caption{Proposed MF-PCAS method}
    \label{alg:MF-PCAS}
    \textbf{Input:} High-fidelity model  $\textit{f}_y : \mathcal{X} \mapsto \mathbf{Y}$, low-fidelity model $\textit{f}_z: \mathcal{X} \mapsto\mathbf{Z}$ \\
    \textbf{Output:}  Predicted high-fidelity output field $\mathbf{\tilde{y}}$
    
    \begin{algorithmic}[1]
        \STATE Reduce output dimensionality using Procrustes manifold alignment $\mathbf{H}, \mathbf{G},\boldsymbol{\Theta}_k = Algorithm 1 (\mathbf{Y},\mathbf{Z})$
        \FOR{$i=1$ to $k$}
            \STATE Find active subspace $\mathbf{{W}}_{AS_{i}} = Algorithm2(\mathbf{H}[i,:],\mathbf{G}[i,:])$
            \STATE Train HK model for $\tilde{h}_i(\mathbf{{W}}_{AS_{i}}\mathbf{x})$
        \ENDFOR
        \STATE For a new design point $\mathbf{x}^*$, predict $$\mathbf{\tilde{y}}=\sum_{i=1}^{k}\boldsymbol{\theta}_i\tilde{h}^{*}_{i}(\mathbf{{W}}_{AS_i}\mathbf{x}^*)$$
        \STATE \textbf{return} $\mathbf{\tilde{y}}$
    \end{algorithmic}
\end{algorithm}

\section{Application Problems}\label{sct: application}
The benefits of our proposed multi-fidelity ROM approach are evaluated using two application problems. The first application problem is the aerodynamic analysis of the RAE~2822 transonic airfoil. For this application problem, the effectiveness of our proposed approach is evaluated using different high and low-fidelity combinations, and parametrizations. The second test problem involves the more complex 3D geometry of the CRM wing. For both test problems, the quantity of interest predicted by the models is the pressure coefficient ($C_p$) of the flow field. The motivation for selecting the $C_p$ field stems from its important role in aircraft design, as it critically influences aerodynamic performance and is necessary for aero-structural design. Both test cases are analyzed with CFD simulations that are performed using the open-source SU2 code~\cite{economon2016su2}.  In the latter parts of this section, performance metrics used to quantify the accuracy of the proposed approach are also introduced. 

\subsection{Transonic Flow over RAE~2822 Airfoil}
\begin{figure}[ht]
    \centering
    \includegraphics[width=1\textwidth]{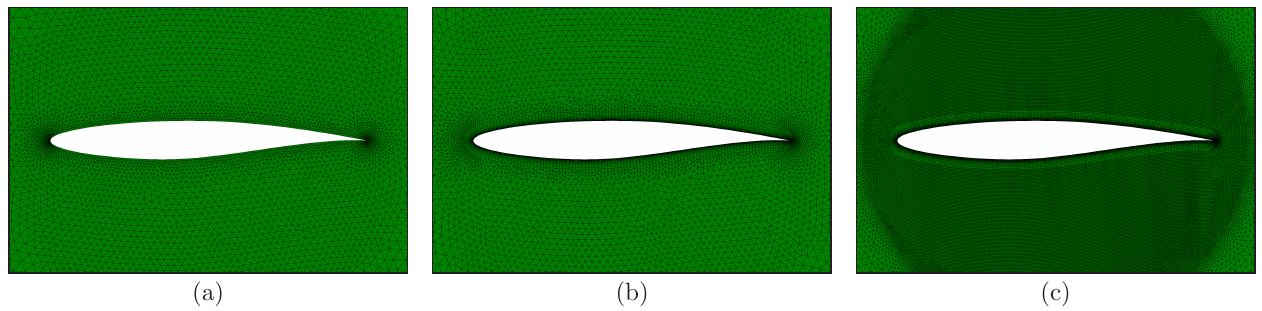}
    \caption{Close-up view of the grids used for RAE~2822 test case: (a) L1 grid (7,485) nodes, (b) L2 grid (19,062 nodes), and (c) L3 grid (96,913 nodes).}
    \label{fig:airfoil-grid}
\end{figure}

Two-dimensional compressible CFD simulations of the RAE~2822 airfoil are performed to evaluate the flow field around an airfoil. RAE~2822 airfoil is the starting geometry for the second benchmark problem for the AIAA aerodynamic design optimization discussion group (ADODG) as it has been used to evaluate many optimization methods~\cite{Lee2015, Poole2015}. Reynolds-averaged-Navier-Stokes (RANS) based simulations are carried out at a free stream flight Mach and Reynolds number of $M_\infty = 0.73$ and $Re_\infty=6.5 \times 10^6$. The Spalart-Allmaras model is used for turbulence modeling~\cite{mufti2019flow,toor2020comparative}. A steady flow solution is obtained using a backward Euler Scheme. High-fidelity flow field results are obtained using an unstructured grid of 96,913 nodes (L3) as shown in Fig.~\ref{fig:airfoil-grid}. Two coarser low-fidelity grids with 19,062 (L2) and 7,485 (L1) nodes are also generated, respectively. The L1 grid is solved using an inviscid solver, which requires a much smaller grid size due to the absence of near-wall refinements normally needed to model boundary layers. Table \ref{table:rae2822-cfd} summarizes the three fidelity levels. The reader is referred to~\cite{Mufti2022AInputs} for the grid convergence study for this test problem.

\begin{table}[ht]
\centering
\caption{Summary of RAE~2822 CFD simulations for different fidelity levels.}\label{table:rae2822-cfd}
\setlength{\tabcolsep}{15pt}
\begin{tabular}{@{}cccc@{}}
\toprule
Level & Num. of Nodes & CFD Solver & Cost/Sample {[}CPU-hr{]} \\ \midrule
L1 & 7,485 & Inviscid & 0.008 \\
L2 & 19,062 & RANS & 0.05 \\
L3 & 96,913 & RANS & 0.329 \\ \bottomrule
\end{tabular}
\multirow{2}{*}{$^*$\footnotesize{CPU times are obtained using Intel Xeon Gold 6226 CPUs}}
\end{table}

\subsection{Transonic Flow Over NASA CRM Wing}
\begin{figure}[ht]
    \centering
    \includegraphics[width=1\textwidth]{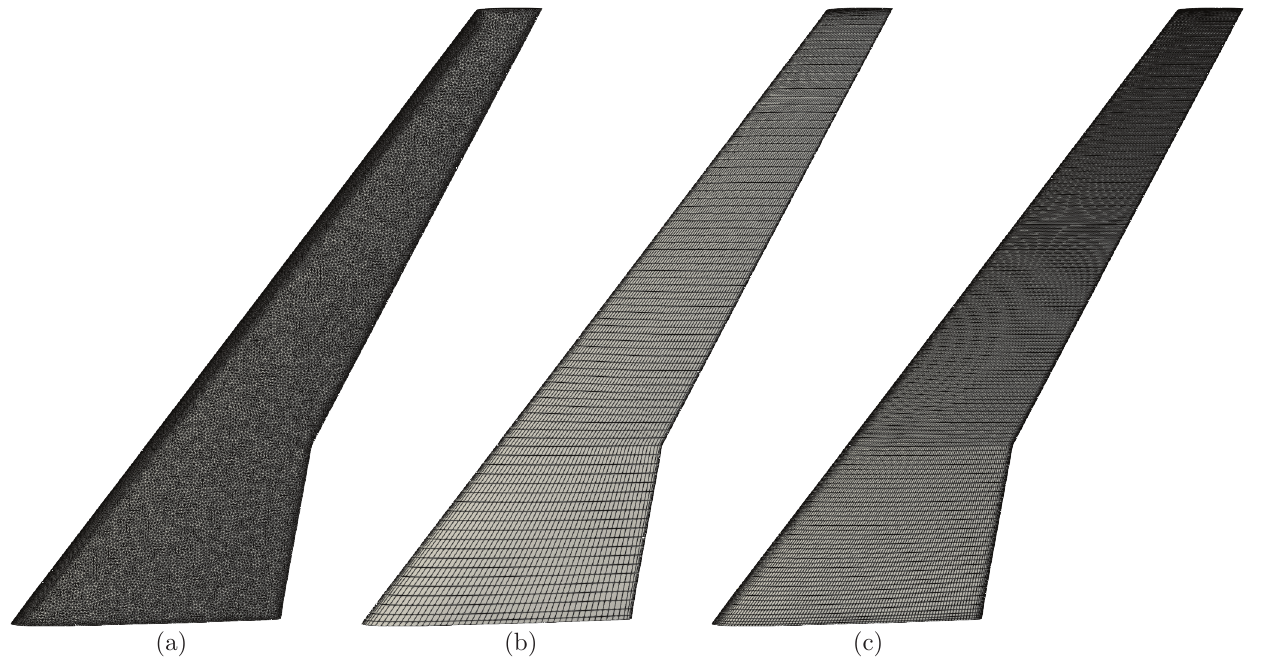}
    \caption
    {Close-up view of grids used for CRM wing test case: (a) L1 unstructured grid (0.26M nodes), (b) L2 structured grid (0.46M nodes), and (c) L3 structured grid (3.7M nodes). Only surface grids are shown in the figure for visualization purposes.}
    \label{fig:CRM_wing}
\end{figure}

A second test case involving the aerodynamic analysis of NASA's common research model (CRM) wing is considered to evaluate our proposed methodology on a problem more representative of an industrial application. The CRM aircraft geometry is also one of the benchmark problems for the ADODG and has been designed to test and develop aerodynamic prediction~\cite{perron2020multi}, and shape optimization methods \cite{Lyu2015,Lee2015}. A CFD setup, similar to RAE 2822 is used. Free stream conditions of $M_\infty=0.85$ and $Re_\infty = 5\times10^6$, similar to one defined by the ADODG are selected. A structured grid of $3.7$ million nodes used for high-fidelity simulations (L3) is shown in Fig.~\ref{fig:CRM_wing}. This grid was also used previously in a multi-level optimization method~\cite{Lyu2015} and has been made public as part of the ADODG benchmark problem. Two coarser grids having $0.46$ million (L2) and $0.26$ million (L1) nodes are generated for low-fidelity solutions. The L1 grid is unstructured and solved using an inviscid solver and hence does not include near-wall refinement. Table. \ref{table:CRM-cfd} gives details for different fidelity grids and their computational time.

\begin{table}[ht]
\centering
\caption{Summary of CRM wing CFD simulations for different fidelity levels.}\label{table:CRM-cfd}
\setlength{\tabcolsep}{15pt}
\begin{tabular}{@{}cccc@{}}
\toprule
Level & Num. of Nodes & CFD Solver & Cost/Sample {[}CPU-hr{]} \\ \midrule
L1 & 251,821 & Inviscid & 1.162 \\
L2 & 464,325 & RANS & 3.384 \\
L3 & 3,659,337 & RANS & 91.20 \\ \bottomrule
\end{tabular}
\multirow{2}{*}{$^*$\footnotesize{CPU times are obtained using Intel Xeon Gold 6226 CPUs}}
\end{table}

\subsection{Geometric Parametrization}
We are interested in predicting field solutions for problems having high-dimensional inputs. Predicting pressure fields for highly parameterized shapes represents a problem having both high-dimensional inputs and outputs. Such problems are frequently encountered in aerodynamic shape optimization studies, where a large number of design variables parameterize aerodynamic geometries. Varying these variables produces complex aerodynamic shapes whose performance is then evaluated and optimized to meet a certain performance target. 

Different shape parametrization methods exist in the literature such as the class shape transformation (CST)~\cite{Kulfan2008}, splines~\cite{Masters2016}, and Hick Hennes bump function~\cite{Hicks1978}. A detailed discussion on different methods used for aerodynamic shape parametrization has been given by~\citet{Masters2017GeometricMethods}. In this study, the free-form deformation (FFD) method is used. The geometry of the RAE~2822 airfoil and CRM wing is enclosed in a rectangular FFD box. The nodes of the FFD box are then displaced causing a change in the shape of the enclosed geometry. The deformation of the FFD node is then propagated into the rest of the grid using a linear elasticity approach. The work of~\citet{Kenway2010} can be referred to for a detailed description of the FFD methodology. 

\begin{figure}[ht]
    \centering
    \includegraphics[width=1\textwidth]{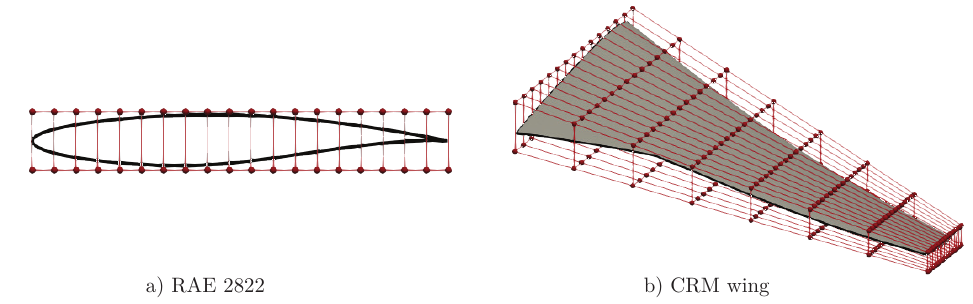}
    \caption{FFD box used for test cases}
    \label{fig:ffd-box}
\end{figure}

For the airfoil test case, four different FFD boxes having a dimension of $7 \times 2$, $12 \times 2$, $22 \times 2$, and $32 \times 2$, are generated as shown in Fig.~\ref{fig:ffd-box}. The four corner FFD nodes are held fixed to prevent the deformation from altering the airfoil angle of attack. The remainder of the nodes is allowed to move only in the vertical direction with a displacement limit of $\pm 0.02$ chord length. Overall, the current setup provides four different scenarios having an input dimension of 10, 20, 40, and 60, respectively. Similarly, an FFD box of $15 \times 8 \times 2$ is generated for the CRM wing case having an input dimension of $240$ as shown in Fig.~\ref{fig:ffd-box}. All the nodes are allowed to move only in the vertical direction with a higher displacement limit of $\pm 0.03$ chord length. 

\subsection{Performance Metrics}
We can quantify the prediction accuracy of our proposed approach by finding the error between the predicted field solution and the actual field solution. For $n_\text{v}$ fields computed for out-of-sample designs, that is, designs not involved in the model's training phase, we calculate the root-mean-squared (RMS) prediction error $\tilde{E}(\mathbf{y})$ as

\begin{equation}\label{eq:pred_error}
    \tilde{E}(\mathbf{y}) = \sqrt{\frac{1}{n_v} \sum_{i=1}^{n_v} \lVert \mathbf{y}^*_i - \mathbf{\Tilde{y}}^*_i \rVert^2}
\end{equation}
where $\mathbf{y}_i^*$ represents the $i$-th out-of-sample solution and $\mathbf{\tilde{y}}_i$ denotes the prediction made by the ROM. 

The framework introduced in this study integrates input and output dimensionality reduction with a regression model. This enables us to split the total prediction error into components of reconstruction and regression errors. The error due to reconstruction, $\tilde{E}_{RC}(\mathbf{y})$, pertains to the error arising from the mapping between the reduced and the original spaces, or in the context of POD, it is the error attributable to the accuracy of the POD modes. The reconstruction error $\tilde{E}_{RC}(\mathbf{y})$ can be expressed as
\begin{equation}\label{eq:recon_error}
\tilde{E}_{RC}(\mathbf{y}) = \sqrt{\frac{1}{n_v} \sum_{i = 1}^{n_\text{v}} \lVert
        (\mathbf{I} - \boldsymbol{\Theta}_k \boldsymbol{\Theta}_k^T)\mathbf{y}_i^*\rVert^2}
\end{equation}

The regression error is defined by the discrepancies within the actual and predicted latent space coordinates due to imperfections in the regression models, whether from inaccuracies in determining the active subspace $\mathbf{\hat{W}}_{AS}$ or the fit of the HK regression model. The regression error $\tilde{E}_{RG}(\mathbf{y})$ is given as
\begin{equation}\label{eq:regr_error}
\tilde{E}_{RG}(\mathbf{y}) = \sqrt{\sum_{i = 1}^{n_\text{v}} \lVert \boldsymbol{\Theta}_k (\boldsymbol{\Theta}_k^T \mathbf{y}_i^* - \mathbf{\tilde{\mathbf{H}}}_i)\rVert^2}
\end{equation}
where $\mathbf{\tilde{\mathbf{H}}}_i$ represents the estimated latent space coordinates for $\mathbf{y}_i^*$. 
Given that the reconstruction and regression errors act as orthogonal components of the total error, we can combine $\tilde{E}_{RC}(\mathbf{y})$ and $\tilde{E}_{RG}(\mathbf{y})$ into the total error $\tilde{E}$, as demonstrated in \cite{perron2020multi}, in the following manner:
\begin{equation}
\tilde{E}(\mathbf{y}) = \sqrt{\tilde{E}_{RC}(\mathbf{y})^2 + \tilde{E}_{RG}(\mathbf{y})^2}
\end{equation}

\section{Results and Discussion}\label{sct: results}
In this section, we present the results of evaluating the predictive performance of the proposed MF-PCAS method for the RAE~2822 airfoil and NASA CRM wing test cases. We compare the accuracy and computational cost of the multi-fidelity MF-PCAS method to the single-fidelity PCAS method to assess the benefits of considering multiple levels of fidelity in creating ROMs for problems with high-dimensional input space. We also examine the error caused by the model-based active subspace and present visual comparisons of the predicted $\tilde{C}_p$ field errors. Finally, we evaluate the performance of the proposed MF-PCAS method as the input dimensionality changes and compare it to the MA-ROM method. To obtain our results, we use datasets containing $2200$ CFD simulations for the RAE~2822 airfoil and $800$ CFD simulations for the CRM wing test case, These datasets are created for each fidelity level described in Tables~\ref{table:rae2822-cfd} and~\ref{table:CRM-cfd}. A subset of $m_1$ training samples is randomly selected from both high- and low-fidelity datasets. A multi-fidelity ratio parameter $\tau = m_2 / m_1$ is used to select an additional $m_2 - m_1$ low-fidelity samples. The MF-PCAS and MA-ROM models are then trained using the selected high- and low-fidelity samples, while the PCAS model is trained using only high-fidelity samples. This process is repeated 20 times or more for each setup, and the mean prediction error is documented for each combination of the size of the training samples and the level of fidelity. In this section, all the ROMs use the fidelity level L3 as the high-fidelity output field of interest and combine it with one of the other fidelity levels to form a multi-fidelity model.

\subsection{Cost vs Accuracy Trade-off}
\begin{figure}[ht]
    \centering
    \includegraphics{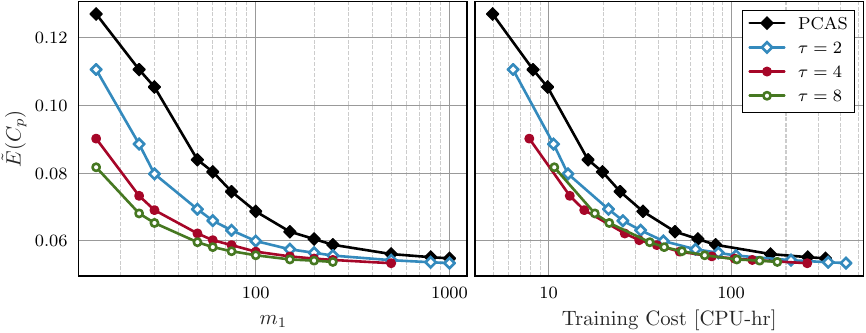}
    \caption{Variation of prediction error $\tilde{E}(C_p)$ with number of high-fidelity samples $m_1$ and training cost for L3 and L2 datasets. Results for RAE~2822 test case with $d = 40$.}
    \label{fig:RAE2822_L3_L2}
\end{figure}

\begin{figure}[ht]
    \centering
    \includegraphics{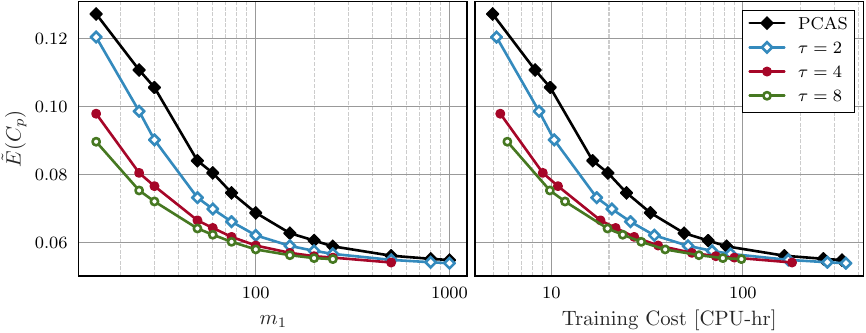}
    \caption{Variation of prediction error $\tilde{E}(C_p)$ with number of high-fidelity samples $m_1$ and training cost for L3 and $L1$ datasets. Results for RAE~2822 test case with $d = 40$}
    \label{fig:RAE2822_L3_L1}
\end{figure}

The objective of the proposed multi-fidelity model is to predict the pressure coefficient ($C_p$) of the high-fidelity flow field L3. The cost vs accuracy trade-off section demonstrates the benefit of combining the expensive-to-obtain L3 data with cheaper data sources. Figures~\ref{fig:RAE2822_L3_L2} and~\ref{fig:RAE2822_L3_L1} present the average prediction errors ($\tilde{E}(C_p)$) of the pressure field for both the single-fidelity PCAS and the multi-fidelity MF-PCAS methods as a function of the number of high-fidelity training samples ($m_1$) and total training cost [CPU-hr]. Figure~\ref{fig:RAE2822_L3_L2} shows the results of combining the fidelity levels L3 + L2, while Fig.~\ref{fig:RAE2822_L3_L1} shows the results of combining L3 + L1 data. Both figures display the errors in predicting the $C_p$ field around the RAE~2822 airfoil with a $d=40$ parametrization.

From the left side of both figures, it is clear that $\tilde{E}(C_p)$ decreases nearly exponentially as the number of training samples increases for all models. The results show that combining L3 with either L2 or L1 data using MF-PCAS consistently reduces $\tilde{E}(C_p)$ for a given $m_1$. The reduction in $\tilde{E}(C_p)$ is higher when using L2 data compared to L1. This is because the L1 data uses a simpler inviscid model and a coarser grid, resulting in a weaker correlation with the high-fidelity data and reduced prediction accuracy of MF-PCAS. Increasing $\tau$ also improves the accuracy of the multi-fidelity method, although with diminishing return as $\tau$ increases to a large value. The MF-PCAS method with both combinations outperforms the single-fidelity PCAS method when $m_1$ is relatively low. As $m_1$ increases to large values, the predictive accuracy of PCAS catches up with MF-PCAS. This suggests that with a large enough training dataset, the single-fidelity model can adequately capture the relationship between inputs and outputs for this test case, rendering the multi-fidelity model superfluous.

The variation of $\tilde{E}(C_p)$ with computational cost to construct the single- and multi-fidelity ROM is shown on the right-hand side of Figs.~\ref{fig:RAE2822_L3_L2} and~\ref{fig:RAE2822_L3_L1}. The computational cost only includes the cost of generating training data as the cost of constructing the ROM is negligible in comparison. It is worth noting that in practical scenarios, the computational cost of generating training data is often limited by a computational budget. The MF-PCAS method provides better prediction accuracy for a fixed computational budget than single-fidelity PCAS. The prediction error decreases when $\tau$ is increased from 2 to 4 for the same training cost. However, beyond a threshold value of $\tau$, adding more low-fidelity data no longer improves the model's training. When the available computational budget is sufficiently large, PCAS and MF-PCAS give similar prediction errors for all values of $\tau$. For such scenarios, the cost of high-fidelity simulations dominates the training cost, and low-fidelity data has minimal impact on both cost and accuracy. Comparing the right-hand sides of Fig.~\ref{fig:RAE2822_L3_L2} and \ref{fig:RAE2822_L3_L1}, we observe that although the combination of L3 + L2 provides a lower prediction error than the combination of L3 + L1 for the same $m_1$ and $\tau$, the cost of generating an L2 sample is $6.25$ times higher than that of an L1 sample. As a result, both combinations, L3 + L2 and L3 + L1 offer a similar cost-accuracy trade-off.

\begin{figure}[ht]
    \centering
    \includegraphics{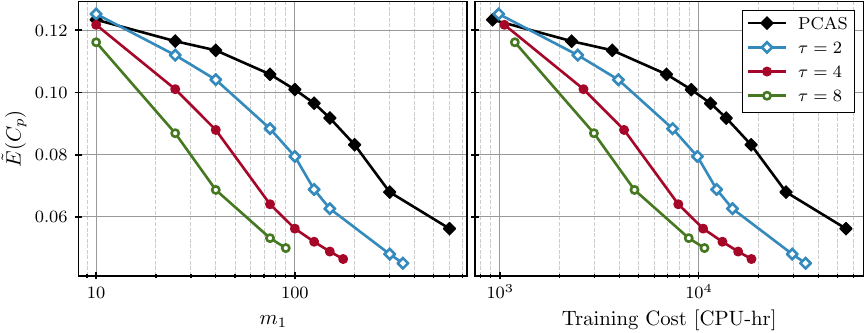}
    \caption{Variation of prediction error $\tilde{E}(C_p)$ with number of high-fidelity samples $m_1$ and training cost for L3 and L2 datasets. Results for CRM wing test case with $d = 240$}
    \label{fig:CRM_L3_L2}
\end{figure}

\begin{figure}[ht]
    \centering
    \includegraphics{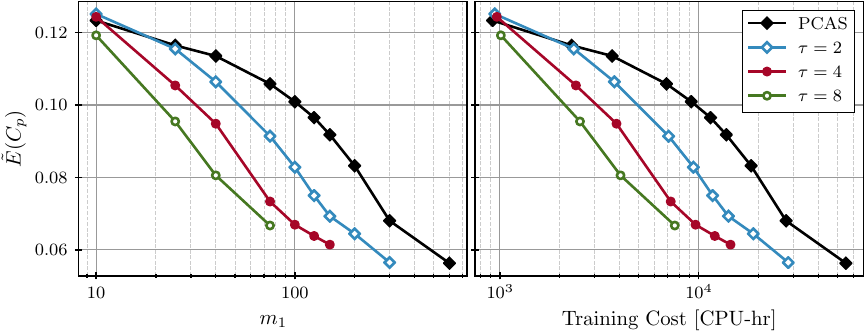}
    \caption{Variation of prediction error $\tilde{E}(C_p)$ with number of high-fidelity samples $m_1$ and training cost for L3 and L1 datasets. Results for CRM wing test case with $d = 240$}
    \label{fig:CRM_L3_L1}
\end{figure}

Figures~\ref{fig:CRM_L3_L2} and~\ref{fig:CRM_L3_L1} show the average prediction errors $\tilde{E}(C_p)$ of the pressure field for PCAS and MF-PCAS methods as a function of $m_1$ and total training cost for CRM wing test case with $d=240$. Unlike the airfoil test case, we only consider the pressure distribution on the wing surface as generally most design applications are focused on predicting aerodynamic and structural loads on the wing surface. We can observe that MF-PCAS using both L3 + L2 and L3 + L1 combinations outperforms single-fidelity PCAS for all values of $m_1$ and total training cost. Even for relatively large values of $m_1$ and training cost, the accuracy of the single-fidelity model fails to match the prediction accuracy of the MF-PCAS method. The 3D wing geometry typically has a more complex flow field and a higher degree of parametrization than the 2D airfoil, making it more difficult to model accurately with single-fidelity PCAS. The use of MF-PCAS with different levels of accuracy allows for a more comprehensive representation of the system. From Figs.~\ref{fig:CRM_L3_L2} and~\ref{fig:CRM_L3_L1} we can observe that increasing $\tau$, for a fixed $m_1$ or computational budget results in a continuous reduction in $\tilde{E}(C_p)$. This shows that adding more low-fidelity data allows the model to learn the complexities of the flow field better.

Similar, to the airfoil test case we see that L3 + L2 and L3 + L1 combinations offer a similar cost vs accuracy trade-off for the CRM wing test case. For the sake of brevity, in the remaining part of this section, we will only show and discuss results for the L3 + L2 fidelity combination for both test cases.


\subsection{Discussion of Prediction Errors}
A detailed breakdown of the prediction errors for the RAE~2822 airfoil and CRM wing test cases is shown in Tables~\ref{tab:error-breakdown_rae2822} and~\ref{tab:error-breakdown_crm}, respectively. Increasing $m_1$ allows for the computation of more accurate and additional POD modes, which results in a quick decrease in the reconstruction error $\tilde{E}_{RC}(C_p)$ for the PCAS model. If an MF-PCAS model is trained with a similar amount of high-fidelity data $m_1$, it will use the same number of POD modes ($k$) to represent the high-fidelity field as discussed in Sec.~\ref{sct: methodology}. As a result, the MF-PCAS will have a comparable $\tilde{E}_{RC}(C_p)$ value to a single-fidelity PCAS model, irrespective of the value of $\tau$ selected. However, using a multi-fidelity approach allows for the accurate computation of the active subspace and the enrichment of the low-dimensional latent space with low-fidelity data. This results in a better fit for the HK regression model and a reduction of $\tilde{E}_{RG}(C_p)$.

\begin{table}[!h]
\caption{Breakdown of the prediction error of the MF-PCAS and PCAS method. Results are for the RAE~2822 test case with $d=40$ and {\normalfont $\text{RIC}=99\%$}. .}
\label{tab:error-breakdown_rae2822}
\centering
\begin{tabular}{@{}ccccccccc@{}}
\toprule
\rowcolor[HTML]{FFFFFF} 
Method & Levels & $m_1$  & $\tau$ & $m_2$  & k  & $\tilde{E}(C_p)$      & $\tilde{E}_{RG}(C_p)$   & $\tilde{E}_{RC}(C_p)$    \\ \midrule
       &        & 50  &   &     & 27 & 0.0839 & 0.0787 & 0.0291  \\
       &        & 100 &   &     & 36 & 0.0686 & 0.0653 & 0.02104 \\
\multirow{-3}{*}{PCAS}    & \multirow{-3}{*}{L3}    & 500                   & \multirow{-3}{*}{-} & \multirow{-3}{*}{-} & 49 & 0.0560 & 0.0541 & 0.0143  \\ \midrule
       &        &     & 2 & 200 & 36 & 0.0598 & 0.0560 & 0.02104 \\
       &        &     & 4 & 400 & 36 & 0.0567 & 0.0527 & 0.02104 \\
\multirow{-3}{*}{MF-PCAS} & \multirow{-3}{*}{L3 + L2} & \multirow{-3}{*}{100} & 8                   & 800                 & 36 & 0.0556 & 0.0515 & 0.02104 \\ \bottomrule
\end{tabular}
\end{table}

From Table~\ref{tab:error-breakdown_crm} we can observe that as $\tau$ increases, the regression error decreases and eventually reaches a magnitude similar to that of the reconstruction error. For such cases, we can further reduce $\tilde{E}_{RC}(C_p)$ by increasing the RIC threshold value. This allows for the inclusion of additional POD modes for high-fidelity field reconstruction. However, the additional POD modes tend to capture more fine-grained details in the high-fidelity field, which can be noise or insignificant variations. Finding a model-based active subspace and fitting a regression model to these modes become challenging and result in an increase in $\tilde{E}_{RG}(C_p)$. Table~\ref{tab:error-breakdown_ric} shows the breakdown of prediction error for both test cases with a RIC = $99.99\%$. Comparing the errors in Table~\ref{tab:error-breakdown_ric} with those in ~Tables~\ref{tab:error-breakdown_rae2822} and~\ref{tab:error-breakdown_crm}, we can observe that for the same $m_1$ and $\tau$, increasing RIC results in an increase in selected POD modes $k$. As a result we see a decrease in $\tilde{E}_{RC}(C_p)$ and an increase in $\tilde{E}_{RG}(C_p)$. The overall prediction error only changes by a relatively small amount.

\begin{table}[!h]
\caption{Breakdown of the prediction error of the MF-PCAS and PCAS method. Results are for the CRM wing test case with $d=240$ and {\normalfont $\text{RIC}=99\%$}.}
\label{tab:error-breakdown_crm}
\centering
\begin{tabular}{@{}ccccccccc@{}}
\toprule
\rowcolor[HTML]{FFFFFF} 
Method & Levels & $m_1$  & $\tau$ & $m_2$  & k  & $\tilde{E}(C_p)$      & $\tilde{E}_{RG}(C_p)$   & $\tilde{E}_{RC}(C_p)$    \\ \midrule
       &        & 25 &   &     & 22 & 0.116  & 0.0995 & 0.0605 \\
       &        & 75 &   &     & 54 & 0.105  & 0.0990 & 0.0372 \\
\multirow{-3}{*}{PCAS}    & \multirow{-3}{*}{L3}    & 500                  & \multirow{-3}{*}{-} & \multirow{-3}{*}{-} & 118 & 0.0679  & 0.0648 & 0.0203 \\ \midrule
       &        &    & 2 & 150 & 54 & 0.0883 & 0.0801 & 0.0372 \\
       &        &    & 4 & 300 & 54 & 0.0640 & 0.0521 & 0.0372 \\
\multirow{-3}{*}{MF-PCAS} & \multirow{-3}{*}{L3 + L2} & \multirow{-3}{*}{75} & 8                   & 600                 & 54  & 0.05321 & 0.0380 & 0.0372 \\ \bottomrule
\end{tabular}
\end{table}

\begin{table}[!h]
\caption{Breakdown of the prediction error of the MF-PCAS method with {\normalfont $\text{RIC}=99.99\%$} . Results are for the RAE~2822 test case with $d=40$ and the CRM Wing Test case with $d=240$.}
\label{tab:error-breakdown_ric}
\centering
\begin{tabular}{@{}ccccccccc@{}}
\toprule
\rowcolor[HTML]{FFFFFF} 
Test Case & Levels & $m_1$  & $\tau$ & $m_2$  & k  & $\tilde{E}(C_p)$      & $\tilde{E}_{RG}(C_p)$   & $\tilde{E}_{RC}(C_p)$    \\ \midrule
          &        &    & 2 & 200 & 91 & 0.0590 & 0.0584 & 0.0130 \\
          &        &    & 4 & 400 & 91 & 0.0564 & 0.0548 & 0.0130 \\
\multirow{-3}{*}{RAE~2822} & \multirow{-3}{*}{L3 + L2} & \multirow{-3}{*}{100} & 8 & 800 & 91 & 0.05521 & 0.05364 & 0.0130 \\ \midrule
          &        &    & 2 & 150 & 75 & 0.0884 & 0.0814 & 0.0343 \\
          &        &    & 4 & 300 & 75 & 0.0634 & 0.0532 & 0.0343 \\
\multirow{-3}{*}{CRM Wing} & \multirow{-3}{*}{L3 + L2}   & \multirow{-3}{*}{75}  & 8 & 600 & 75 & 0.05212 & 0.03918 & 0.0343 \\ \bottomrule
\end{tabular}
\end{table}

\begin{figure}[!h]
    \centering
    \includegraphics[scale =1.2]{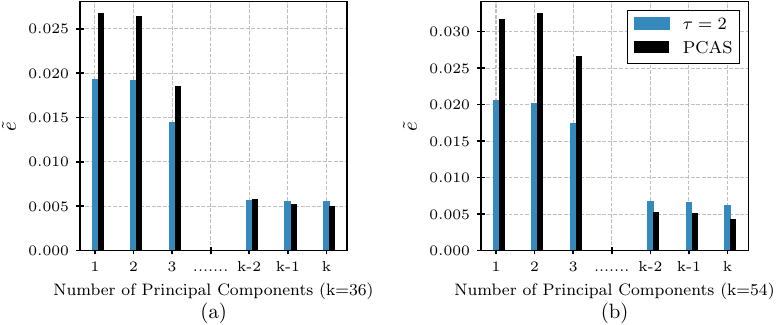}
    \caption{Regression error in predicting each POD mode with {\normalfont $\text{RIC}=99\%$}.(a) RAE~2822 test case with $m_1 = 100$ and $d=40$ (b) CRM wing test case with $m_1 =75$ and $d=240$}
    \label{fig:reg_error_modes}
\end{figure}

The error due to the multi-fidelity model-based active subspace is included in the regression error of the model. We can write the total regression error in Eq.~\eqref{eq:regr_error} as the root sum of squares of the errors in predicting each POD mode
\begin{equation}\label{eq:regr_error_2}
\tilde{E}_{RG}(C_{p}) = \sqrt{\sum_{j = 1}^{k}\sum_{i = 1}^{n_\text{v}} \lVert \boldsymbol{\theta}_j (\boldsymbol{\theta}_j^T \mathbf{y}_i^* - \mathbf{\tilde{h}}_j)\rVert^2}
\end{equation}
\begin{equation}\label{eq:regr_error_3}
\tilde{E}_{RG}(C_{p}) = \sqrt{\tilde{e}^{2}_{1} + \tilde{e}^{2}_{2} + \dots + \tilde{e}^{2}_{k-1} + \tilde{e}^{2}_{k}}
\end{equation}
where $\tilde{e} = \sum_{i = 1}^{n_\text{v}} \lVert \boldsymbol{\theta} (\boldsymbol{\theta}^T \mathbf{y}_i^* - \mathbf{\tilde{h}})\rVert$ is the error in predicting each POD mode coefficient. From Fig.~\ref{fig:reg_error_modes} we can see that for both test cases, the MF-PCAS model that uses a multi-fidelity active subspace to reduce input space dimensionality has a lower prediction error for the first few dominant POD. For the latter noisy modes, both the single-fidelity and multi-fidelity methods fail to discover a good active subspace and produce an inaccurate regression model. However, the contribution of these latter modes is less significant than the first few dominant modes in overall prediction error, which lessens their impact on the ROM.

\subsection{Visualization of Error in Flow Field Prediction}
\begin{figure}[!h]
    \centering
    \includegraphics[scale=0.75]{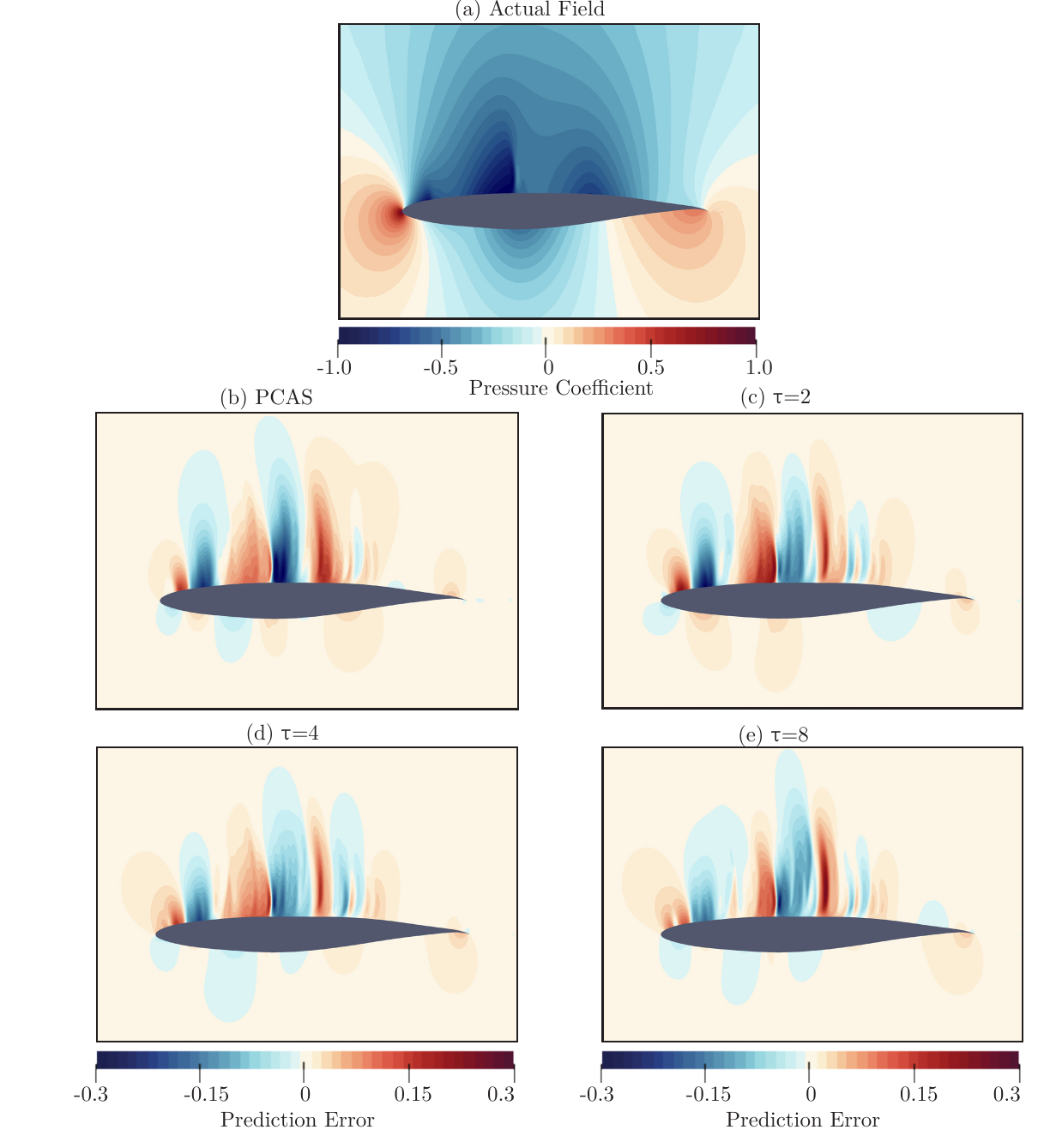}
    \caption{Prediction error ($C_{p} - \tilde{C}_{p}$) in the flow field. Results for RAE~2822 airfoil test case with $m_1 = 100$ and $d = 40$}
    \label{fig:RAE2822_field}
\end{figure}

The figures presented in this subsection provide a visual representation of flow field errors and complement the previous discussions and explanations of errors observed in the experiments. Figure~\ref{fig:RAE2822_field} illustrates the CFD pressure coefficient ($C_p$) field and prediction error ($C_p - \tilde{C}_p$) field for the RAE~2822 airfoil test case. The selected airfoil design exhibits complex flow features on the top surface of the airfoil, including multiple shocks and separation bubbles. The MF-PCAS method predicts the flow field accurately, and the prediction error is almost zero for most of the flow domain. However, spurious oscillations on the top surface of the airfoil are observed, although the magnitude of the error is lower than that of the single-fidelity prediction. Additionally, Fig.~\ref{fig:RAE2822_field} confirms the previous observation that beyond a certain threshold, adding further low-fidelity data has no significant effect on the prediction accuracy of the MF-PCAS method for the airfoil test case. We can observe that increasing the value of $\tau$ from 4 to 8 does not significantly impact the error contours.

\begin{figure}[!h]
    \centering
    \includegraphics[scale=0.75]{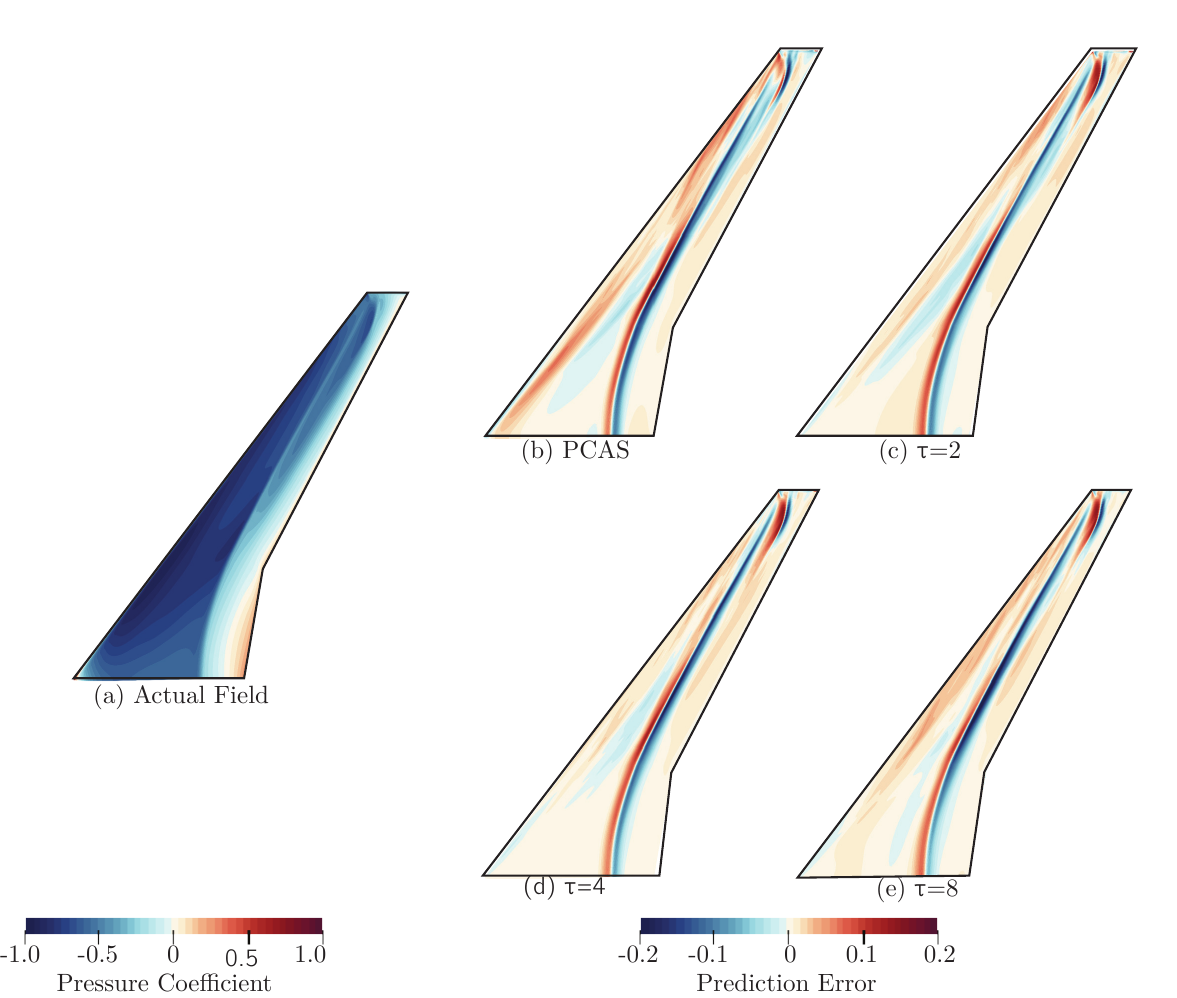}
    \caption{Prediction error ($C_{p} - \tilde{C}_{p}$) in the flow field. Results for CRM wing test case with $m_1 = 75$ and $d = 240$}
    \label{fig:CRM_field}
\end{figure}

In Fig.~\ref{fig:CRM_field}, we examine the pressure and predicted error fields for the top surface of the CRM wing test case, where abrupt pressure changes and complex flow phenomena typically occur. The comparison of the fields in Fig.~\ref{fig:CRM_field} indicates that most of the prediction error for all models is concentrated near a strong shock. Both the single- and multi-fidelity methods utilize POD for output dimensionality reduction. POD is a linear method and assumes that a linear combination of variables can capture the dominant patterns in the field. However, shockwaves represent a highly non-linear phenomenon, and POD cannot capture their sudden and rapid changes in the field. Nevertheless, the MF-PCAS method performs better in capturing the shockwave when compared to its single-fidelity counterpart. Moreover, adding more low-fidelity data by increasing $\tau$ results in a further reduction in prediction error near the shock region, which is consistent with the previously observed trend.

\subsection{Effect of Input Dimensionality}
To evaluate the effect of input dimensionality on the predictive performance of the MF-PCAS method, we analyze four different airfoil test cases with input dimensions of 10, 20, 40, and 60, respectively. The test cases are generated using different FFD parameterizations, as described in Sec.~\ref{sct: application}. We compare the results with those of the manifold-aligned ROM (MA-ROM) method, which carries out output dimension reduction using Procrustes manifold alignment but develops the ROM over the original high-dimensional input space. For both the MF-PCAS and MA-ROM methods, we use an HK model with automatic relevance determination squared-exponential kernel for the prediction of latent POD modes. 

\begin{figure}[!h]
    \centering
    \includegraphics{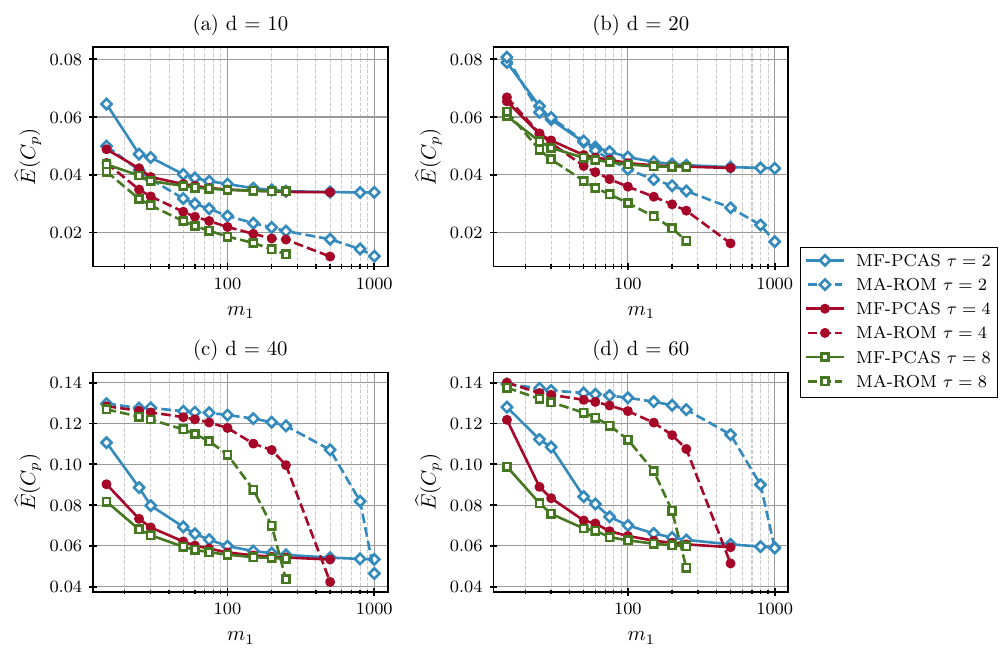}
    \caption{Effect of Input Dimensionality on total prediction error $\tilde{E}(C_p)$}
    \label{fig:dim_study}
\end{figure}

Figure~\ref{fig:dim_study} shows the variation of $\tilde{E}(C_p)$ with $m_1$ and $\tau$ for different input dimensions, $d$. As the number of input dimensions increases, the total prediction error increases for both the MA-ROM and MF-PCAS methods. An increasing number of input dimensions results in airfoil shapes with complex flow fields that are difficult to predict. We observe that when $d=10$, the MA-ROM method outperformed the MF-PCAS method for all values of $m_1$ and $\tau$. For the same number of high-fidelity training samples, MA-ROM and MF-PCAS are expected to have similar reconstruction errors. When input dimensionality is low, the MA-ROM method is not expected to suffer from the curse of dimensionality and would have a low regression error. On the contrary, finding a multi-fidelity model-based active subspace to reduce input dimensionality actually introduces additional errors and increases the regression error for MF-PCAS. When the input dimensionality is increased to 20, the MF-PCAS method accuracy catches up with that of the MA-ROM method for low $m_1$ values. As $m_1$ increases to a higher value, the MA-ROM method's accuracy improves significantly as sufficient training data becomes available for the model to overcome dimensionality issues.

When the input dimensionality is increased to a sufficiently large value, we see a clear advantage of using MF-PCAS over MA-ROM. Fig.~\ref{fig:dim_study} shows that for $d=40$ and 60, the MF-PCAS method consistently performs better than the MA-ROM method. For high-dimensional input problems, a large number of training samples are needed to fit an accurate prediction model due to the curse of dimensionality. Reducing input dimensionality by finding a multi-fidelity model-based active subspace helps to alleviate the effects of the curse of dimensionality and reduces the regression error and total prediction error $\tilde{E}(C_p)$. The performance of the MA-ROM model only catches up with MF-PCAS for very large $m_1$. However, generating such a large amount of high-fidelity data often becomes computationally infeasible.

Table~\ref{tab:dim_stuy_diff} shows the total prediction error for the MA-ROM and MF-PCAS models trained using the same number of high- and low-fidelity samples when the input dimension is 40. The computational cost of training both models is the same as differences in the model fitting cost were negligible compared to the cost of generating training data. From the table, we can observe that, for the same training sample size and cost, MF-PCAS offers around 50\% better prediction accuracy than MA-ROM. This prediction accuracy advantage slightly decreases to 46\% when we add more low-fidelity training data by increasing $\tau$ from 2 to 8.  Despite this decrease, this experiment clearly shows the advantage of using the MF-PCAS method over the MA-ROM when the input dimensionality is large.

\begin{table}[!h]
\caption{Comparison of total prediction error $\tilde{E}(C_p)$ for MF-PCAS and MA-ROM. Results are for the RAE~2822 test case with $d=40$.}
\label{tab:dim_stuy_diff}
\centering
\begin{tabular}{@{}cccccccccc@{}}
\toprule
 &  &  &  &  &  &  & \multicolumn{3}{c}{ $\tilde{E}(C_p)$} \\ \cmidrule(l){8-10} 
\multirow{-2}{*}{Test Case} & \multirow{-2}{*}{Levels} & \multirow{-2}{*}{$m_1$} & \multirow{-2}{*}{T} & \multirow{-2}{*}{$m_1$} & \multirow{-2}{*}{k} & \multirow{-2}{*}{Cost {[}CPU-hr{]}} & \cellcolor[HTML]{FFFFFF}MF-PCAS & \cellcolor[HTML]{FFFFFF}MA-ROM & \cellcolor[HTML]{FFFFFF}$\%$ Change \\ \midrule
 &  &  & 2 & 200 & 36 & 42.6 & 0.0598 & 0.124 & 51.77\% \\
 &  &  & 4 & 400 & 36 & 52.3 & 0.0567 & 0.1179 & 51.90\% \\
\multirow{-3}{*}{\begin{tabular}[c]{@{}c@{}}RAE~2822 \end{tabular}} & \multirow{-3}{*}{L3+L2} & \multirow{-3}{*}{100} & 8 & 800 & 36 & 71.7 & 0.0556 & 0.10475 & 46.92\% \\ \bottomrule
\end{tabular}
\end{table}

\section{Conclusion}
This work has successfully demonstrated the efficacy of a multi-fidelity, parametric, and non-intrusive ROM framework for high-dimensional input spaces. Our approach integrates machine learning concepts of output field dimension reduction via POD, Procrustes manifold alignment, input space reduction using a gradient-free model-based active subspace, and multi-fidelity regression. We evaluated our approach using a 2D RAE 2822 airfoil and a 3D NASA CRM wing. These evaluations have shown our methodology to be superior to the single-fidelity PCAS method, particularly in achieving a favorable cost and accuracy trade-off. The results indicate that our approach offers comparable predictive accuracy at a reduced computational cost and outperforms the manifold-aligned ROM (MA-ROM) by 50\% in scenarios with large input dimensions.

While the current work has focused on aerodynamic design cases, the framework is versatile enough to be applied to any surrogate modeling problem that involves high-dimensional inputs and outputs and multiple data sources. This universality underscores the potential for broad application across various domains of engineering design.

A notable limitation identified in this study is the use of a linear POD technique for output dimension reduction and linear Procrustes manifold alignment, which can restrict the model’s ability to accurately predict non-linear phenomena such as shock waves, as observed in our results. To address this limitation, future work will concentrate on incorporating non-linear deep learning-based dimensionality reduction and manifold alignment techniques. This advancement is expected to enhance the predictive performance of our model, especially in scenarios involving complex non-linearities like shock waves.

%\section{Acknowledgements and Disclaimer}

\bibliography{references,references_mendeley}

\begin{thebibliography}{74}
\newcommand{\enquote}[1]{``#1''}
\providecommand{\natexlab}[1]{#1}
\providecommand{\url}[1]{\texttt{#1}}
\providecommand{\urlprefix}{URL }
\expandafter\ifx\csname urlstyle\endcsname\relax
  \providecommand{\doi}[1]{\discretionary{}{}{}https://doi.org/#1}\else
  \providecommand{\doi}[1]{\discretionary{}{}{}\urlstyle{rm}\url{https://doi.org/#1}}\fi

\bibitem[{Benner et~al.(2015)Benner, Gugercin, and Willcox}]{benner2015survey}
Benner, P., Gugercin, S., and Willcox, K., \enquote{A survey of projection-based model reduction methods for parametric dynamical systems,} \emph{SIAM review}, Vol.~57, No.~4, 2015, pp. 483--531.

\bibitem[{Yondo et~al.(2018)Yondo, Andr{\'e}s, and Valero}]{yondo2018review}
Yondo, R., Andr{\'e}s, E., and Valero, E., \enquote{A review on design of experiments and surrogate models in aircraft real-time and many-query aerodynamic analyses,} \emph{Progress in aerospace sciences}, Vol.~96, 2018, pp. 23--61.

\bibitem[{Forrester and Keane(2009)}]{Forrester2009}
Forrester, A.~I., and Keane, A.~J., \enquote{{Recent advances in surrogate-based optimization},} \emph{Progress in Aerospace Sciences}, Vol.~45, No. 1-3, 2009, pp. 50--79.
\newblock \doi{10.1016/j.paerosci.2008.11.001}.

\bibitem[{Xiao et~al.(2015)Xiao, Fang, Buchan, Pain, Navon, and Muggeridge}]{Xiao2015Non-intrusiveEquations}
Xiao, D., Fang, F., Buchan, A.~G., Pain, C.~C., Navon, I.~M., and Muggeridge, A., \enquote{{Non-intrusive reduced order modelling of the Navier-Stokes equations},} \emph{Computer Methods in Applied Mechanics and Engineering}, Vol. 293, 2015, pp. 522--541.
\newblock \doi{10.1016/j.cma.2015.05.015}.

\bibitem[{Lu et~al.(2019)Lu, Jin, Chen, Yang, Hou, Zhang, Li, and Fu}]{lu2019review}
Lu, K., Jin, Y., Chen, Y., Yang, Y., Hou, L., Zhang, Z., Li, Z., and Fu, C., \enquote{Review for order reduction based on proper orthogonal decomposition and outlooks of applications in mechanical systems,} \emph{Mechanical Systems and Signal Processing}, Vol. 123, 2019, pp. 264--297.

\bibitem[{Fossati(2015)}]{Fossati2015EvaluationMethodology}
Fossati, M., \enquote{{Evaluation of aerodynamic loads via reduced-order methodology},} \emph{AIAA Journal}, Vol.~53, No.~8, 2015, pp. 2389--2405.
\newblock \doi{10.2514/1.J053755}.

\bibitem[{Decker et~al.(2020)Decker, Schwartz, and Mavris}]{decker2020dimensionality}
Decker, K., Schwartz, H.~D., and Mavris, D., \enquote{Dimensionality reduction techniques applied to the design of hypersonic aerial systems,} \emph{AIAA Aviation 2020 Forum}, 2020, p. 3003.

\bibitem[{Rajaram et~al.(2020{\natexlab{a}})Rajaram, Perron, Puranik, and Mavris}]{rajaram2020randomized}
Rajaram, D., Perron, C., Puranik, T.~G., and Mavris, D.~N., \enquote{Randomized algorithms for non-intrusive parametric reduced order modeling,} \emph{AIAA Journal}, Vol.~58, No.~12, 2020{\natexlab{a}}, pp. 5389--5407.

\bibitem[{Behere et~al.(2021)Behere, Rajaram, Puranik, Kirby, and Mavris}]{Behere2021ReducedEstimation}
Behere, A., Rajaram, D., Puranik, T.~G., Kirby, M., and Mavris, D.~N., \enquote{{Reduced order modeling methods for aviation noise estimation},} \emph{Sustainability (Switzerland)}, Vol.~13, No.~3, 2021, pp. 1--19.
\newblock \doi{10.3390/su13031120}.

\bibitem[{Chen and Thuerey(2023)}]{Chen2023TowardsAerofoils}
Chen, L.~W., and Thuerey, N., \enquote{{Towards high-accuracy deep learning inference of compressible flows over aerofoils},} \emph{Computers {\&} Fluids}, Vol. 250, 2023, p. 105707.
\newblock \doi{10.1016/J.COMPFLUID.2022.105707}.

\bibitem[{Kashefi et~al.(2021)Kashefi, Rempe, and Guibas}]{Kashefi2021AGeometries}
Kashefi, A., Rempe, D., and Guibas, L.~J., \enquote{{A point-cloud deep learning framework for prediction of fluid flow fields on irregular geometries},} \emph{Physics of Fluids}, Vol.~33, No.~2, 2021.
\newblock \doi{10.1063/5.0033376}.

\bibitem[{Deng et~al.(2023)Deng, Wang, Liu, Xie, Li, Zhang, Jia, Zhang, Wang, and Dong}]{Deng2023PredictionStrategies}
Deng, Z., Wang, J., Liu, H., Xie, H., Li, B.~K., Zhang, M., Jia, T., Zhang, Y., Wang, Z., and Dong, B., \enquote{{Prediction of transonic flow over supercritical airfoils using geometric-encoding and deep-learning strategies},} \emph{Physics of Fluids}, Vol.~35, No.~7, 2023.
\newblock \doi{10.1063/5.0155383}.

\bibitem[{Mufti et~al.(2024)Mufti, Bhaduri, Ghosh, Wang, and Mavris}]{mufti2024shock}
Mufti, B., Bhaduri, A., Ghosh, S., Wang, L., and Mavris, D.~N., \enquote{Shock wave prediction in transonic flow fields using domain-informed probabilistic deep learning,} \emph{Physics of Fluids}, Vol.~36, No.~1, 2024.

\bibitem[{Koch et~al.(1999)Koch, Simpson, Allen, and Mistree}]{Koch1999}
Koch, P.~N., Simpson, T.~W., Allen, J.~K., and Mistree, F., \enquote{{Statistical Approximations for Multidisciplinary Design Optimization: The Problem of Size},} \emph{Journal of Aircraft}, Vol.~36, No.~1, 1999, pp. 275--286.
\newblock \doi{10.2514/2.2435}, \urlprefix\url{https://arc.aiaa.org/doi/10.2514/2.2435}.

\bibitem[{Kumari and Jayaram(2017)}]{Kumari2017}
Kumari, S., and Jayaram, B., \enquote{{Measuring Concentration of Distances—An Effective and Efficient Empirical Index},} \emph{IEEE Transactions on Knowledge and Data Engineering}, Vol.~29, No.~2, 2017, pp. 373--386.
\newblock \doi{10.1109/TKDE.2016.2622270}.

\bibitem[{Brunton et~al.(2019)Brunton, Noack, and Koumoutsakos}]{Brunton2019AnnualMechanics}
Brunton, S.~L., Noack, B.~R., and Koumoutsakos, P., \enquote{{Annual Review of Fluid Mechanics Machine Learning for Fluid Mechanics},} \emph{Annu. Rev. Fluid Mech. 2020}, Vol.~52, 2019, pp. 477--508.
\newblock \doi{10.1146/annurev-fluid-010719}, \urlprefix\url{https://doi.org/10.1146/annurev-fluid-010719-}.

\bibitem[{Dietrich et~al.(2018)Dietrich, K{\"u}nzner, Neckel, K{\"o}ster, and Bungartz}]{dietrich2018fast}
Dietrich, F., K{\"u}nzner, F., Neckel, T., K{\"o}ster, G., and Bungartz, H.-J., \enquote{Fast and flexible uncertainty quantification through a data-driven surrogate model,} \emph{International Journal for Uncertainty Quantification}, Vol.~8, No.~2, 2018.

\bibitem[{Saltelli et~al.(2008)Saltelli, Ratto, Andres, Campolongo, Cariboni, Gatelli, Saisana, and Tarantola}]{saltelli2008global}
Saltelli, A., Ratto, M., Andres, T., Campolongo, F., Cariboni, J., Gatelli, D., Saisana, M., and Tarantola, S., \emph{Global sensitivity analysis: the primer}, John Wiley \& Sons, 2008.

\bibitem[{Neal(1998)}]{neal1998assessing}
Neal, R.~M., \enquote{Assessing relevance determination methods using DELVE,} \emph{Nato Asi Series F Computer And Systems Sciences}, Vol. 168, 1998, pp. 97--132.

\bibitem[{Bouhlel et~al.(2016)Bouhlel, Bartoli, Otsmane, and Morlier}]{Bouhlel2016}
Bouhlel, M.~A., Bartoli, N., Otsmane, A., and Morlier, J., \enquote{An Improved Approach for Estimating the Hyperparameters of the Kriging Model for High-Dimensional Problems through the Partial Least Squares Method,} \emph{Mathematical Problems in Engineering}, Vol. 2016, 2016.
\newblock \doi{10.1155/2016/6723410}.

\bibitem[{Constantine et~al.(2014)Constantine, Dow, and Wang}]{Constantine2014}
Constantine, P.~G., Dow, E., and Wang, Q., \enquote{{Active Subspace Methods in Theory and Practice: Applications to Kriging Surfaces},} \emph{SIAM Journal on Scientific Computing}, Vol.~36, No.~4, 2014, pp. A1500--A1524.
\newblock \doi{10.1137/130916138}, \urlprefix\url{http://arxiv.org/abs/1304.2070 http://dx.doi.org/10.1137/130916138 http://epubs.siam.org/doi/10.1137/130916138}.

\bibitem[{Berguin and Mavris(2015)}]{Berguin2015DimensionalityGradient}
Berguin, S.~H., and Mavris, D.~N., \enquote{{Dimensionality reduction using principal component analysis applied to the gradient},} \emph{AIAA Journal}, Vol.~53, American Institute of Aeronautics and Astronautics Inc., 2015, pp. 1078--1090.
\newblock \doi{10.2514/1.J053372}.

\bibitem[{Berguin et~al.(2015)Berguin, Rancourt, and Mavris}]{Berguin2015MethodAnalyses}
Berguin, S.~H., Rancourt, D., and Mavris, D.~N., \enquote{{Method to facilitate high-dimensional design space exploration using computationally expensive analyses},} \emph{AIAA Journal}, Vol.~53, American Institute of Aeronautics and Astronautics Inc., 2015, pp. 3752--3765.
\newblock \doi{10.2514/1.J054035}.

\bibitem[{Jiang et~al.(2020)Jiang, Hu, Liu, Liang, and Wang}]{Jiang2020AQuantification}
Jiang, X., Hu, X., Liu, G., Liang, X., and Wang, R., \enquote{{A generalized active subspace for dimension reduction in mixed aleatory-epistemic uncertainty quantification},} \emph{Computer Methods in Applied Mechanics and Engineering}, Vol. 370, 2020.
\newblock \doi{10.1016/j.cma.2020.113240}.

\bibitem[{Lukaczyk et~al.(2014)Lukaczyk, Constantine, Palacios, and Alonso}]{Lukaczyk2014}
Lukaczyk, T.~W., Constantine, P., Palacios, F., and Alonso, J.~J., \enquote{Active subspaces for shape optimization,} \emph{10th AIAA multidisciplinary design optimization conference}, 2014, p. 1171.

\bibitem[{Li et~al.(2019)Li, Cai, and Qu}]{Li2019Surrogate-basedMethod}
Li, J., Cai, J., and Qu, K., \enquote{{Surrogate-based aerodynamic shape optimization with the active subspace method},} \emph{Structural and Multidisciplinary Optimization}, Vol.~59, No.~2, 2019, pp. 403--419.
\newblock \doi{10.1007/s00158-018-2073-5}.

\bibitem[{Tezzele et~al.(2019)Tezzele, Demo, and Rozza}]{tezzele2019shape}
Tezzele, M., Demo, N., and Rozza, G., \enquote{Shape optimization through proper orthogonal decomposition with interpolation and dynamic mode decomposition enhanced by active subspaces,} \emph{arXiv preprint arXiv:1905.05483}, 2019.

\bibitem[{Wei et~al.(2023)Wei, An, Zhang, Zhou, and Ren}]{wei2023exploiting}
Wei, J., An, J., Zhang, Q., Zhou, H., and Ren, Z., \enquote{Exploiting Active Subspaces for Geometric Optimization of Cavity-Stabilized Supersonic Flames,} \emph{AIAA Journal}, 2023, pp. 1--12.

\bibitem[{Constantine et~al.(2015)Constantine, Emory, Larsson, and Iaccarino}]{Constantine2015}
Constantine, P.~G., Emory, M., Larsson, J., and Iaccarino, G., \enquote{Exploiting active subspaces to quantify uncertainty in the numerical simulation of the HyShot II scramjet,} \emph{Journal of Computational Physics}, Vol. 302, 2015, pp. 1--20.
\newblock \doi{https://doi.org/10.1016/j.jcp.2015.09.001}, \urlprefix\url{https://www.sciencedirect.com/science/article/pii/S002199911500580X}.

\bibitem[{Khatamsaz et~al.(2021)Khatamsaz, Molkeri, Couperthwaite, James, Arr{\'{o}}yave, Srivastava, and Allaire}]{Khatamsaz2021AdaptiveDesign}
Khatamsaz, D., Molkeri, A., Couperthwaite, R., James, J., Arr{\'{o}}yave, R., Srivastava, A., and Allaire, D., \enquote{{Adaptive active subspace-based efficient multifidelity materials design},} \emph{Materials and Design}, Vol. 209, 2021.
\newblock \doi{10.1016/j.matdes.2021.110001}.

\bibitem[{Constantine and Doostan(2017)}]{Constantine2017Time-dependentModel}
Constantine, P.~G., and Doostan, A., \enquote{{Time-dependent global sensitivity analysis with active subspaces for a lithium ion battery model},} \emph{Statistical Analysis and Data Mining}, Vol.~10, No.~5, 2017, pp. 243--262.
\newblock \doi{10.1002/sam.11347}.

\bibitem[{Lam et~al.(2020)Lam, Zahm, Marzouk, and Willcox}]{Lam2020MultifidelitySubspaces}
Lam, R.~R., Zahm, O., Marzouk, Y.~M., and Willcox, K.~E., \enquote{{Multifidelity dimension reduction via active subspaces},} \emph{SIAM Journal on Scientific Computing}, Vol.~42, No.~2, 2020, pp. A929--A956.
\newblock \doi{10.1137/18M1214123}.

\bibitem[{Romor et~al.(2023)Romor, Tezzele, Mrosek, Othmer, and Rozza}]{romor2023multi}
Romor, F., Tezzele, M., Mrosek, M., Othmer, C., and Rozza, G., \enquote{Multi-fidelity data fusion through parameter space reduction with applications to automotive engineering,} \emph{International Journal for Numerical Methods in Engineering}, Vol. 124, No.~23, 2023, pp. 5293--5311.

\bibitem[{Mufti et~al.(2022)Mufti, Chen, Perron, and Mavris}]{Mufti2022AInputs}
Mufti, B., Chen, M., Perron, C., and Mavris, D.~N., \enquote{{A Multi-Fidelity Approximation of the Active Subspace Method for Surrogate Models with High-Dimensional Inputs},} \emph{AIAA AVIATION 2022 Forum}, American Institute of Aeronautics and Astronautics Inc, AIAA, 2022.
\newblock \doi{10.2514/6.2022-3488}.

\bibitem[{Tripathy et~al.(2016)Tripathy, Bilionis, and Gonzalez}]{Tripathy2016GaussianPropagation}
Tripathy, R., Bilionis, I., and Gonzalez, M., \enquote{{Gaussian processes with built-in dimensionality reduction: Applications to high-dimensional uncertainty propagation},} \emph{Journal of Computational Physics}, Vol. 321, 2016, pp. 191--223.
\newblock \doi{10.1016/j.jcp.2016.05.039}.

\bibitem[{Tsilifis et~al.(2021)Tsilifis, Pandita, Ghosh, Andreoli, Vandeputte, and Wang}]{Tsilifis2021BayesianProcesses}
Tsilifis, P., Pandita, P., Ghosh, S., Andreoli, V., Vandeputte, T., and Wang, L., \enquote{{Bayesian learning of orthogonal embeddings for multi-fidelity Gaussian Processes},} \emph{Computer Methods in Applied Mechanics and Engineering}, Vol. 386, 2021.
\newblock \doi{10.1016/j.cma.2021.114147}.

\bibitem[{Gautier et~al.(2022)Gautier, Pandita, Ghosh, and Mavris}]{gautier2022fully}
Gautier, R., Pandita, P., Ghosh, S., and Mavris, D., \enquote{A Fully Bayesian Gradient-Free Supervised Dimension Reduction Method using Gaussian Processes,} \emph{International Journal for Uncertainty Quantification}, Vol.~12, No.~2, 2022.

\bibitem[{Vohra et~al.(2020)Vohra, Nath, Mahadevan, and Tina~Lee}]{Vohra2020FastManufacturing}
Vohra, M., Nath, P., Mahadevan, S., and Tina~Lee, Y.~T., \enquote{{Fast surrogate modeling using dimensionality reduction in model inputs and field output: Application to additive manufacturing},} \emph{Reliability Engineering and System Safety}, Vol. 201, 2020.
\newblock \doi{10.1016/j.ress.2020.106986}.

\bibitem[{Guo et~al.(2023)Guo, Mahadevan, Matsumoto, Taba, and Watanabe}]{guo2023investigation}
Guo, Y., Mahadevan, S., Matsumoto, S., Taba, S., and Watanabe, D., \enquote{Investigation of Surrogate Modeling Options with High-Dimensional Input and Output,} \emph{AIAA Journal}, Vol.~61, No.~3, 2023, pp. 1334--1348.

\bibitem[{Demo et~al.(2019)Demo, Tezzele, and Rozza}]{demo2019non}
Demo, N., Tezzele, M., and Rozza, G., \enquote{A non-intrusive approach for the reconstruction of POD modal coefficients through active subspaces,} \emph{Comptes Rendus M{\'e}canique}, Vol. 347, No.~11, 2019, pp. 873--881.

\bibitem[{Ji et~al.(2022)Ji, Xiao, and Zhan}]{Ji2022HighTechnique}
Ji, Y., Xiao, N.~C., and Zhan, H., \enquote{{High dimensional reliability analysis based on combinations of adaptive Kriging and dimension reduction technique},} \emph{Quality and Reliability Engineering International}, Vol.~38, No.~5, 2022, pp. 2566--2585.
\newblock \doi{10.1002/qre.3091}.

\bibitem[{Rajaram et~al.(2020{\natexlab{b}})Rajaram, Gautier, Perron, Pinon-Fischer, and Mavris}]{Rajaram2020}
Rajaram, D., Gautier, R.~H., Perron, C., Pinon-Fischer, O.~J., and Mavris, D., \enquote{Non-intrusive parametric reduced order models with high-dimensional inputs via gradient-free active subspace,} \emph{AIAA AVIATION 2020 FORUM}, 2020{\natexlab{b}}, p. 3184.

\bibitem[{O'Leary-Roseberry et~al.(2022)O'Leary-Roseberry, Villa, Chen, and Ghattas}]{OLeary-Roseberry2022Derivative-informedPDEs}
O'Leary-Roseberry, T., Villa, U., Chen, P., and Ghattas, O., \enquote{{Derivative-informed projected neural networks for high-dimensional parametric maps governed by PDEs},} \emph{Computer Methods in Applied Mechanics and Engineering}, Vol. 388, 2022.
\newblock \doi{10.1016/j.cma.2021.114199}.

\bibitem[{Forrester et~al.(2007)Forrester, S{\'{o}}bester, and Keane}]{Forrester2007Multi-fidelityModelling}
Forrester, A.~I., S{\'{o}}bester, A., and Keane, A.~J., \enquote{{Multi-fidelity optimization via surrogate modelling},} \emph{Proceedings of the Royal Society A: Mathematical, Physical and Engineering Sciences}, Vol. 463, No. 2088, 2007.
\newblock \doi{10.1098/rspa.2007.1900}.

\bibitem[{Peherstorfer et~al.(2018)Peherstorfer, Willcox, and Gunzburger}]{peherstorfer2018survey}
Peherstorfer, B., Willcox, K., and Gunzburger, M., \enquote{Survey of multifidelity methods in uncertainty propagation, inference, and optimization,} \emph{Siam Review}, Vol.~60, No.~3, 2018, pp. 550--591.

\bibitem[{Bertram et~al.(2018)Bertram, Othmery, and Zimmermannz}]{Bertram2018TowardsModeling}
Bertram, A., Othmery, C., and Zimmermannz, R., \enquote{{Towards real-time vehicle aerodynamic design via multi-fidelity data-driven reduced order modeling},} \emph{AIAA/ASCE/AHS/ASC Structures, Structural Dynamics, and Materials Conference, 2018}, Vol.~0, American Institute of Aeronautics and Astronautics Inc, AIAA, 2018.
\newblock \doi{10.2514/6.2018-0916}.

\bibitem[{Wang et~al.(2011)Wang, Krafft, Mahadevan, Ma, and Fu}]{wang2011manifold}
Wang, C., Krafft, P., Mahadevan, S., Ma, Y., and Fu, Y., \enquote{Manifold alignment,} \emph{Manifold Learning: Theory and Applications}, Vol. 510, 2011.

\bibitem[{Perron(2020)}]{perron2020multi}
Perron, C., \enquote{Multi-fidelity reduced-order modeling applied to fields with inconsistent representations,} Ph.D. thesis, Atlanta: Georgia Institute of Technology, 2020.

\bibitem[{Perron et~al.(2021)Perron, Rajaram, and Mavris}]{Perron2021Multi-fidelityAlignment}
Perron, C., Rajaram, D., and Mavris, D.~N., \enquote{{Multi-fidelity non-intrusive reduced-order modelling based on manifold alignment},} \emph{Proceedings of the Royal Society A: Mathematical, Physical and Engineering Sciences}, Vol. 477, No. 2253, 2021.
\newblock \doi{10.1098/rspa.2021.0495}.

\bibitem[{Perron et~al.(2022)Perron, Sarojini, Rajaram, Corman, and Mavris}]{Perron2022ManifoldAnalysis}
Perron, C., Sarojini, D., Rajaram, D., Corman, J., and Mavris, D., \enquote{{Manifold alignment-based multi-fidelity reduced-order modeling applied to structural analysis},} \emph{Structural and Multidisciplinary Optimization}, Vol.~65, No.~8, 2022.
\newblock \doi{10.1007/s00158-022-03274-1}.

\bibitem[{Decker et~al.(2023)Decker, Iyengar, Rajaram, Perron, and Mavris}]{Decker2023ManifoldModeling}
Decker, K., Iyengar, N., Rajaram, D., Perron, C., and Mavris, D., \enquote{{Manifold Alignment-Based Nonintrusive and Nonlinear Multifidelity Reduced-Order Modeling},} \emph{AIAA Journal}, Vol.~61, No.~1, 2023, pp. 454--474.
\newblock \doi{10.2514/1.J061720}.

\bibitem[{Li et~al.(2022)Li, Du, and Martins}]{li2022machine}
Li, J., Du, X., and Martins, J.~R., \enquote{Machine learning in aerodynamic shape optimization,} \emph{Progress in Aerospace Sciences}, Vol. 134, 2022, p. 100849.

\bibitem[{Wang and Mahadevan(2009)}]{wang2009general}
Wang, C., and Mahadevan, S., \enquote{A general framework for manifold alignment,} \emph{2009 AAAI Fall Symposium Series}, 2009.

\bibitem[{Guerrero et~al.(2014)Guerrero, Ledig, and Rueckert}]{guerrero2014manifold}
Guerrero, R., Ledig, C., and Rueckert, D., \enquote{Manifold alignment and transfer learning for classification of Alzheimer’s disease,} \emph{International Workshop on Machine Learning in Medical Imaging}, Springer, 2014, pp. 77--84.

\bibitem[{Bousmalis et~al.(2016)Bousmalis, Trigeorgis, Silberman, Krishnan, and Erhan}]{bousmalis2016domain}
Bousmalis, K., Trigeorgis, G., Silberman, N., Krishnan, D., and Erhan, D., \enquote{Domain separation networks,} \emph{Advances in neural information processing systems}, Vol.~29, 2016.

\bibitem[{Wang and Mahadevan(2008)}]{wang2008manifold}
Wang, C., and Mahadevan, S., \enquote{Manifold alignment using procrustes analysis,} \emph{Proceedings of the 25th international conference on Machine learning}, 2008, pp. 1120--1127.

\bibitem[{Pinnau(2008)}]{pinnau2008model}
Pinnau, R., \enquote{Model reduction via proper orthogonal decomposition,} \emph{Model order reduction: theory, research aspects and applications}, Springer, 2008, pp. 95--109.

\bibitem[{Gower(2010)}]{gower2010procrustes}
Gower, J.~C., \enquote{Procrustes methods,} \emph{Wiley Interdisciplinary Reviews: Computational Statistics}, Vol.~2, No.~4, 2010, pp. 503--508.

\bibitem[{Choi et~al.(2009)Choi, Alonso, and Kroo}]{Choi2009Two-levelJets}
Choi, S., Alonso, J.~J., and Kroo, H.~M., \enquote{{Two-level multifidelity design optimization studies for supersonic jets},} \emph{Journal of Aircraft}, Vol.~46, No.~3, 2009, pp. 776--790.
\newblock \doi{10.2514/1.34362}.

\bibitem[{Han et~al.(2012)Han, {Zimmermann}, and G{\"{o}}rtz}]{Han2012AlternativeModeling}
Han, Z.~H., {Zimmermann}, and G{\"{o}}rtz, S., \enquote{{Alternative cokriging model for variable-fidelity surrogate modeling},} \emph{AIAA Journal}, Vol.~50, No.~5, 2012, pp. 1205--1210.
\newblock \doi{10.2514/1.J051243}.

\bibitem[{Han and Görtz(2012)}]{Han2012}
Han, Z.~H., and Görtz, S., \enquote{Hierarchical kriging model for variable-fidelity surrogate modeling,} \emph{AIAA Journal}, Vol.~50, 2012, pp. 1885--1896.
\newblock \doi{10.2514/1.J051354}.

\bibitem[{Mufti et~al.(2023)Mufti, Perron, Gautier, and Mavris}]{mufti2023design}
Mufti, B., Perron, C., Gautier, R., and Mavris, D.~N., \enquote{Design Space Reduction using Multi-Fidelity Model-Based Active Subspaces,} \emph{AIAA AVIATION 2023 Forum}, 2023, p. 3592.

\bibitem[{Han and G{\"{o}}rtz(2012)}]{Han2012HierarchicalModeling}
Han, Z.~H., and G{\"{o}}rtz, S., \enquote{{Hierarchical kriging model for variable-fidelity surrogate modeling},} \emph{AIAA Journal}, Vol.~50, No.~9, 2012, pp. 1885--1896.
\newblock \doi{10.2514/1.J051354}.

\bibitem[{Economon et~al.(2016)Economon, Palacios, Copeland, Lukaczyk, and Alonso}]{economon2016su2}
Economon, T.~D., Palacios, F., Copeland, S.~R., Lukaczyk, T.~W., and Alonso, J.~J., \enquote{SU2: An open-source suite for multiphysics simulation and design,} \emph{AIAA Journal}, Vol.~54, No.~3, 2016, pp. 828--846.

\bibitem[{Lee et~al.(2015)Lee, Koo, Telidetzki, Buckley, Gagnon, and Zingg}]{Lee2015}
Lee, C., Koo, D., Telidetzki, K., Buckley, H., Gagnon, H., and Zingg, D.~W., \enquote{Aerodynamic shape optimization of benchmark problems using jetstream,} \emph{53rd AIAA Aerospace Sciences Meeting}, 2015, p. 0262.

\bibitem[{Poole et~al.(2015)Poole, Allen, and Rendall}]{Poole2015}
Poole, D.~J., Allen, C.~B., and Rendall, T., \enquote{Control point-based aerodynamic shape optimization applied to AIAA ADODG test cases,} \emph{53rd AIAA Aerospace Sciences Meeting}, 2015, p. 1947.

\bibitem[{Mufti et~al.(2019)Mufti, Khan, Masud, and Toor}]{mufti2019flow}
Mufti, B., Khan, T., Masud, J., and Toor, Z., \enquote{Flow field analysis of a diverterless supersonic inlet using embedded les methodology,} \emph{2019 IEEE 10th International Conference on Mechanical and Aerospace Engineering (ICMAE)}, IEEE, 2019, pp. 79--84.

\bibitem[{Toor et~al.(2020)Toor, Masud, Irfan, Mufti, and Khan}]{toor2020comparative}
Toor, Z., Masud, J., Irfan, T., Mufti, B., and Khan, O., \enquote{Comparative Analysis of Aerodynamic Characteristics of a Transport Aircraft and its AWACS Variant,} \emph{AIAA Scitech 2020 Forum}, 2020, p. 2227.

\bibitem[{Lyu et~al.(2015)Lyu, Kenway, and Martins}]{Lyu2015}
Lyu, Z., Kenway, G. K.~W., and Martins, J. R. R.~A., \enquote{{Aerodynamic Shape Optimization Investigations of the Common Research Model Wing Benchmark},} \emph{AIAA Journal}, Vol.~53, No.~4, 2015, pp. 968--985.
\newblock \doi{10.2514/1.J053318}, \urlprefix\url{https://arc.aiaa.org/doi/10.2514/1.J053318}.

\bibitem[{Kulfan(2008)}]{Kulfan2008}
Kulfan, B.~M., \enquote{{Universal Parametric Geometry Representation Method},} \emph{Journal of Aircraft}, Vol.~45, No.~1, 2008, pp. 142--158.
\newblock \doi{10.2514/1.29958}.

\bibitem[{Masters et~al.(2016)Masters, Poole, Taylor, Rendall, and Allen}]{Masters2016}
Masters, D.~A., Poole, D.~J., Taylor, N.~J., Rendall, T., and Allen, C.~B., \enquote{Impact of shape parameterisation on aerodynamic optimisation of benchmark problem,} \emph{54th AIAA Aerospace Sciences Meeting}, 2016, p. 1544.

\bibitem[{Hicks and Henne(1978)}]{Hicks1978}
Hicks, R.~M., and Henne, P.~A., \enquote{{Wing Design By Numerical Optimization},} \emph{Journal of Aircraft}, Vol.~15, No.~7, 1978, pp. 407--412.
\newblock \doi{10.2514/3.58379}.

\bibitem[{Masters et~al.(2017)Masters, Taylor, Rendall, Allen, and Poole}]{Masters2017GeometricMethods}
Masters, D.~A., Taylor, N.~J., Rendall, T.~C., Allen, C.~B., and Poole, D.~J., \enquote{{Geometric comparison of aerofoil shape parameterization methods},} \emph{AIAA Journal}, Vol.~55, No.~5, 2017, pp. 1575--1589.
\newblock \doi{10.2514/1.J054943}.

\bibitem[{Kenway et~al.(2010)Kenway, Kennedy, and Martins}]{Kenway2010}
Kenway, G., Kennedy, G., and Martins, J., \enquote{{A CAD-Free Approach to High-Fidelity Aerostructural Optimization},} \emph{13th AIAA/ISSMO Multidisciplinary Analysis Optimization Conference}, American Institute of Aeronautics and Astronautics, Reston, Virigina, 2010, pp. 1--18.
\newblock \doi{10.2514/6.2010-9231}, \urlprefix\url{http://arc.aiaa.org/doi/10.2514/6.2010-9231}.

\end{thebibliography}

\end{document}